\documentclass[lettersize,journal]{IEEEtran}

\IEEEoverridecommandlockouts                              

\usepackage{subcaption}
\usepackage[utf8]{inputenc}
\usepackage[english]{babel}

\usepackage{booktabs}
\usepackage{amsmath,amsthm}
\usepackage{multicol}
\usepackage{mathtools}
\usepackage[table]{xcolor}
\usepackage[verbose]{wrapfig}
\usepackage{url}
\usepackage{svg}
\usepackage{caption}
\usepackage{threeparttable}
\usepackage{graphicx}
\usepackage{array}
\usepackage{hhline}
\usepackage{breqn}
\usepackage{soul}
\usepackage{amssymb}
\usepackage{lipsum}
\usepackage{enumitem}
\usepackage{multirow}
\usepackage{tcolorbox}

\usepackage{arydshln}

\newtheorem{definition}{Definition}

\definecolor{readgreen}{RGB}{0,150,0}

\title{\LARGE \bf
Generalized Multi-hop Downstream Traffic Pressure for Heterogeneous Perimeter Control
}

\author{Xiaocan Li$^{1}$, Xiaoyu Wang$^{2}$, Ilia Smirnov$^{2}$, Scott Sanner$^{1}$, and Baher Abdulhai$^{2}$

\thanks{$^{1}$Xiaocan Li and Scott Sanner are with the Department of Mechanical \& Industrial Engineering, University of Toronto, Canada
        {\tt\small \{hsiaotsan.li, ssanner\}@mail.utoronto.ca}}
        
\thanks{$^{2}$Xiaoyu Wang, Ilia Smirnov, and Baher Abdulhai are with the Department of Civil \& Mineral Engineering, University of Toronto, Canada
        {\tt\small cnxiaoyu.wang@mail.utoronto.ca, \{ilia.smirnov, baher.abdulhai\}@utoronto.ca}}
}

\begin{document}

\maketitle

\begin{abstract}
Perimeter control (PC) prevents loss of traffic network capacity due to congestion in urban areas. Homogeneous PC allows all access points to a protected region to have identical permitted inflow. However, homogeneous PC performs poorly when the congestion in the protected region is heterogeneous (e.g., imbalanced demand) since the homogeneous PC does not consider specific traffic conditions around each perimeter intersection. When the protected region has spatially heterogeneous congestion, one needs to modulate the perimeter inflow rate to be higher near low-density regions and vice versa for high-density regions. A naïve approach is to leverage 1-hop traffic pressure to measure traffic condition around perimeter intersections, but such metric is too spatially myopic for PC. To address this issue, we formulate multi-hop downstream pressure grounded on Markov chain theory, which ``looks deeper'' into the protected region beyond perimeter intersections. In addition, we formulate a two-stage hierarchical control scheme that can leverage this novel multi-hop pressure to redistribute the total permitted inflow provided by a pre-trained deep reinforcement learning homogeneous control policy. Experimental results show that our heterogeneous PC approaches leveraging multi-hop pressure significantly outperform homogeneous PC in scenarios where the origin-destination flows are highly imbalanced with high spatial heterogeneity. Moveover, our approach is shown to be robust against turning ratio uncertainties by a sensitivity analysis.
\end{abstract}


\section{Introduction}\label{s:intro}
\IEEEPARstart{P}{erimeter} control regulates transfer flows among protected regions by limiting the permitted inflow from the feeder links to protected regions, ensuring that each protected region's vehicle accumulation does not exceed a critical value that would negatively impact traffic productivity in the protected regions. Perimeter control usually relies on Macroscopic Fundamental Diagrams (MFDs), which describe the relationship between traffic productivity and congestion levels and are widely used in numerous studies \cite{daganzo2007urban, daganzo2008analytical, daganzo2011macroscopic}. For MFD-based perimeter control, control parameters design relies on the function of MFDs and critical value of vehicle accumulation \cite{keyvan2012exploiting, keyvan2015controller, keyvan2021optimizing}. However, the accuracy of MFDs is severely affected by the homogeneity of congestion distribution within each region.
Lower homogeneity (higher heterogeneity) results in lower accuracy of MFDs \cite{mazloumian2010spatial}, and consequently adversely affects the performance of MFD-based perimeter controllers.

\begin{figure*}[!t]
    \centering 
    \begin{subfigure}[t]{0.7\columnwidth}
        \centering
        \includegraphics[width=\linewidth]{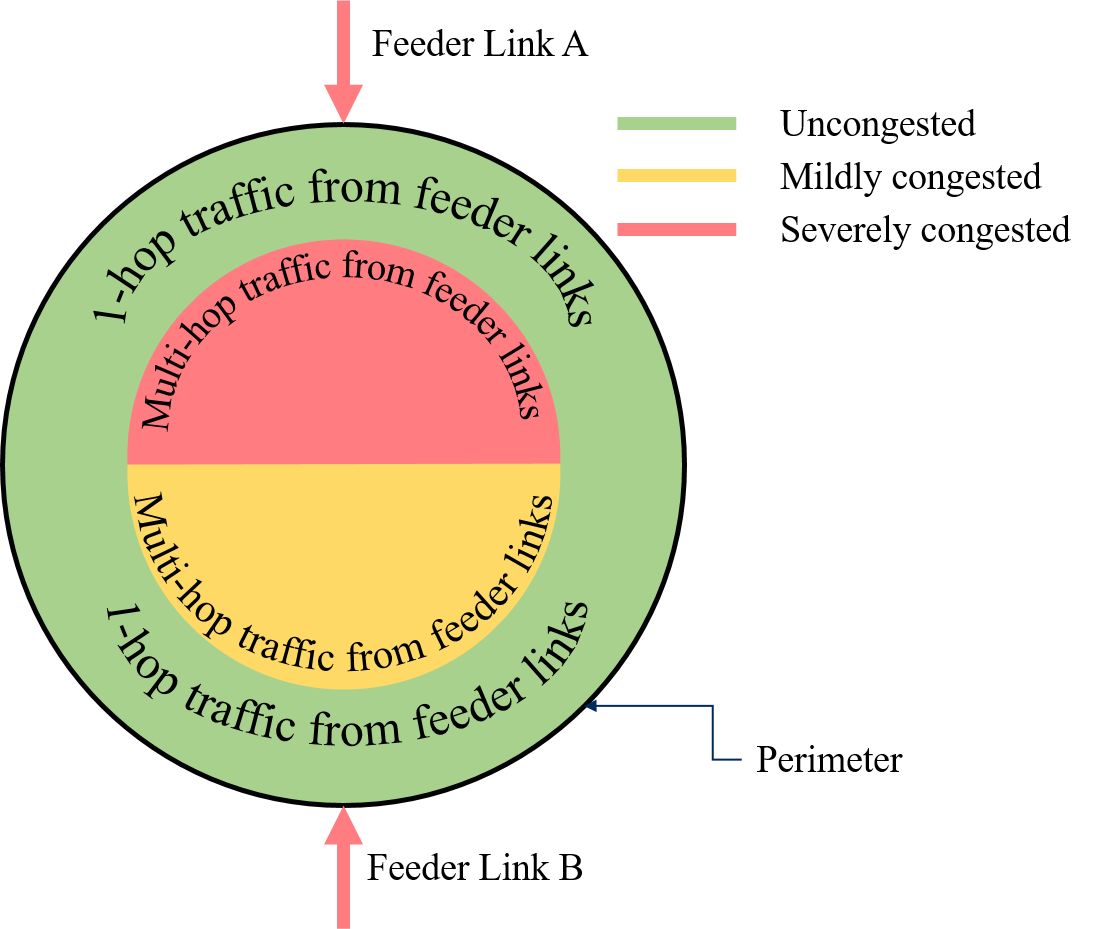}
        \caption{A generic scheme.}
        \label{fig:generic-scheme}
    \end{subfigure}
    \hspace{20pt} 
    \begin{subfigure}[t]{0.56\columnwidth}
        \centering
        \includegraphics[width=\linewidth]{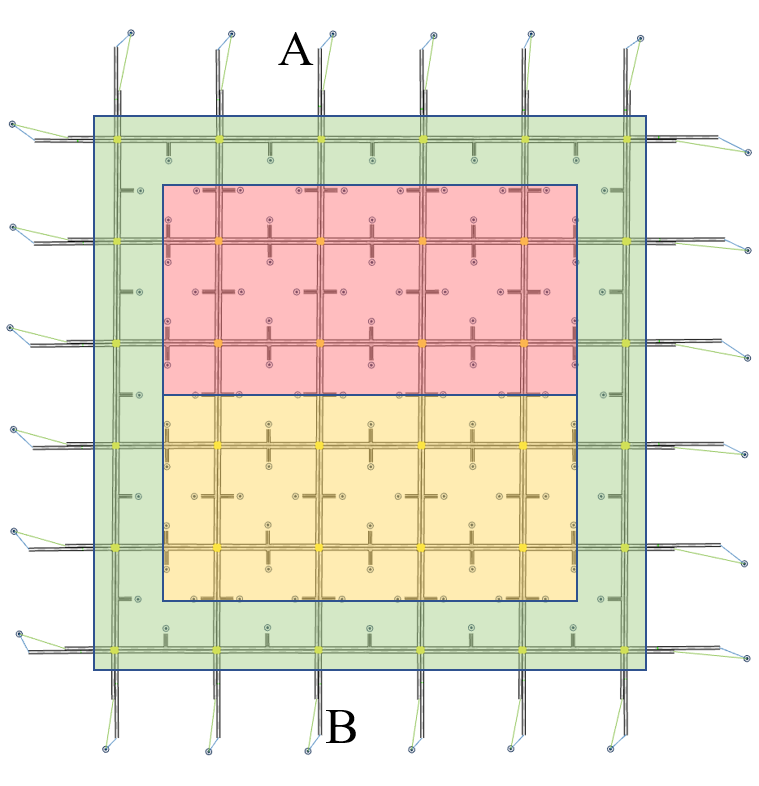}
        \caption{A specific scheme.}
        \label{fig:specific-scheme}
    \end{subfigure}
    \caption{Generic and specific schemes showing 1-hop pressure is myopic for perimeter control and the need for multi-hop pressure.}
    \vspace{-5pt}
    \label{fig:scheme-1-hop-pressure-limitation}
\end{figure*}

As a concrete example of the limitations of the homogeneity assumption of MFDs, Figure~\ref{fig:generic-scheme} demonstrates a protected region with internally imbalanced congestion.  In this case, choosing a homogeneous inflow around the perimeter may oversaturate the upper subregion while under-utilizing traffic capacity in the lower subregion.

In contrast to MFDs that model aggregated congestion metrics at the region and network level, there is also a class of metrics aiming to measure congestion at the local intersection level.  Specifically, the notion of 1-hop traffic pressure quantifies the traffic difference between an upstream link and its immediate downstream links, which is inspired by related work in communication networks~\cite{tassiulas1990stability, neely2003dynamic, georgiadis2006resource}.  Formally, the mathematical definition of 1-hop pressure is in Eq. (\ref{eq:1-hop-pressure-generic}), where $S(l)$ is a traffic statistic (e.g., queue density) for link $l$, $\mathcal{N}(l,1)$ is the set of 1-hop downstream links for link $l$, and $T_{lj}$ is the turning ratio from link $l$ to $j$:
\begin{align}\label{eq:1-hop-pressure-generic}
p(l, 1) = S(l) - \sum_{j \in \mathcal{N}(l,1)} T_{lj} S(j)
\end{align}

From the perspective of 1-hop pressure, we can analyze an intersection-level view of congestion in Figure~\ref{fig:specific-scheme}. The pressure at each perimeter intersection provides a local indication of where to provide perimeter inflow to minimize congestion.  In this case, feeder links at the lower perimeter (e.g., feeder link B) have higher pressure than feeder links at the upper subregion (e.g., feeder link A), making it more sensible to permit larger inflows in the lower subregion.

While modulating perimeter inflows according to 1-hop pressure seems to provide a clear solution to the drawbacks of MFD-based homogeneous PC as described above, the existing 1-hop pressure fails to differentiate feeder link A and B, and therefore cannot improve the performance of heterogeneous PC. This creates a need for a metric with \textit{intermediate spatial granularity} that accounts for the gap between the intersection-level metrics (1-hop pressure) and network-level metrics (MFDs) to effectively reflect the nature of heterogeneous PC. 

To address the limitations of existing traffic metrics, we formalize a generalized concept of multi-hop downstream pressure to incorporate a farsighted assessment of protected region congestion. The primary objective is to extend the existing myopic 1-hop traffic pressure metric into a \textit{multi-hop downstream} traffic pressure metric. Informed by this metric, a heterogeneous perimeter controller can redistribute permitted inflow more effectively based on the specific conditions of each feeder link. Empirically, we demonstrate that our heterogeneous control methodology, which leverages multi-hop pressure, reduces collective travel time compared to the homogeneous approach, heterogeneous PC with myopic pressures, and other baseline heterogeneous PC methods under imbalanced protected region congestion. The contributions of our work are outlined below:
\begin{enumerate}
    \item We provide a mathematical formalization of the multi-hop downstream traffic pressure based on Markov chain theory. This generalization allows us to adjust how far we can reach downstream links, providing a customizable spatial granularity of metrics that bridges the gap between the overly extensive scope of MFDs and the very limited scope of the traditional traffic pressure metric;
    \item We conduct empirical evaluations with a simple controller leveraging multi-hop pressure for heterogeneous perimeter control under scenarios of various demand heterogeneity;
    \item We perform sensitivity analysis that highlights the robustness of our approach against the uncertainties in turning ratio estimations.
\end{enumerate}

\section{Literature Review}

\subsection{Perimeter control}\label{sec:lit-review-pc}
Perimeter control manages inter-regional transfer flows by gating the feeder links of protected regions, ensuring that each region's vehicle count does not surpass a critical value that would negatively impact traffic efficiency \cite{keyvan2012exploiting, keyvan2015controller, keyvan2019traffic, NI2019358, liu2024nmp}. Urban traffic networks frequently become oversaturated during rush hour, necessitating perimeter control. The delineation of a congested area is called perimeter identification, which can be executed statically \cite{saeedmanesh2016clustering} or dynamically \cite{saeedmanesh2017dynamic, li2021perimeter}. Once the perimeter is identified, perimeter control operates at the network level, and is particularly effective in oversaturated situations where adaptive traffic signal control alone is insufficient \cite{keyvan2019traffic}.

Perimeter control can also be classified as either homogeneous or heterogeneous based on whether the inflow rate is uniformly applied across all access points (homogeneous) or adapted based on local conditions (heterogeneous):

\paragraph{Homogeneous Perimeter Control} 
Homogeneous PC treats the protected region as a single entity without accounting for local variations around access points. This approach is particularly compatible with macroscopic traffic models (synonyms: aggregated models, system dynamics), where individual feeder links are not distinguished. The approaches for homogeneous PC include the following key categories:
\begin{itemize}
    \item \textit{Feedback Control}: The inflow rate is adjusted to maintain traffic density near a target point, or ``set-point'' based on the region's MFD. PI controllers are commonly used for this purpose, setting a critical density to optimize traffic flow for the entire region \cite{keyvan2012exploiting, keyvan2015controller, keyvan2019traffic, haddad2014robust}.
    \item \textit{Model Predictive Control (MPC)}: MPC techniques \cite{mpc20132012geroliminisoptimalcontrol, mpc2013geroliminis} have been adapted for homogeneous PC where the control actions are computed by solving the optimization problem over a finite time horizon considering aggregated models, and objectives are related to regional traffic efficiency. 
    \item \textit{Learning-based Control}: Instead of designing controllers based on traffic models, learning-based approaches learn a control policy through the interaction with traffic environments. \cite{ren2020data} estimated pseudo-Jacobian matrix for control from historical data. \cite{su-neuro-dynamic2020} proposed to approximate the value function with neural networks for the optimal value function in HJB equations. With deep reinforcement learning (RL), \cite{zhou2021modelfree} formulated a model-free RL controller for a single protected region PC, then extended to multiple protected regions \cite{zhou2023scalable}, but both works are evaluated on aggregated traffic models. Instead, \cite{li2023perimeter, su2023hierarchical} trained RL on actual traffic networks with microscopic simulation for homogeneous PC, and \cite{li2023perimeter} visualized the contribution of each state variable to the policy.
\end{itemize}


\paragraph{Heterogeneous Perimeter Control} Heterogeneous PC allocates different inflows to each access point based on local traffic conditions. Although works \cite{ramezani2015dynamics, jiang2023hybrid} studied demand heterogeneity with MPC and \cite{li2021perimeter} with linear quadratic regulator, they only tested on aggregated models instead of actual traffic networks. Therefore, these works do not fall under the category of heterogeneous PC with inflow redistribution around perimeter access points. 

The architecture of heterogeneous PC is usually two-stage: the first stage calculates total permitted inflow for the protected region based on homogeneous PC. At the second stage, the total permitted inflow is redistributed based on local traffic conditions. This approach enables more adaptive control and is particularly beneficial in microscopic simulations that capture detailed network features. The redistribution policies at the second stage can be categorized into the following:
\begin{itemize}
    \item \textit{Rule-based Redistribution Policies}: These policies distribute total inflow based on simple rules, such as proportional redistribution to the static attribute -- saturation flows of feeder links \cite{keyvan2012exploiting, keyvan2015controller, ABOUDOLAS2013265,  kouvelas2017enhancing}. This provides a straightforward approach but may not adapt well to dynamic traffic conditions. Instead, \cite{keyvan2015concentric} integrated the real-time queue length of feeder links with the rule: if low demand or spillback occur, then unused inflows are reallocated to other feeder links. \cite{yu2025perimeter}  modifies the Q-value of signal phases on perimeter intersections according to the total travel time in protected regions.
    \item \textit{Optimization-based Redistribution Policies}: Inflow redistribution is framed as an knapsack problem, e.g., \cite{keyvan2021optimizing} used quadratic programming to redistribute the total permitted inflow from the first-stage PI controller, where the objective is to balance relative queues or delays on feeder links. \cite{tsitsokas2023two} leveraged queues on feeder links to weight the importance of phases for perimeter intersections on the quadratic objective function; \cite{iordanidou2017feedback} allocates inflow metering rates based on upstream delays.
    \item \textit{Learning-based Redistribution Policies}: Recent studies have applied RL and graph neural networks (GNNs) to heterogeneous control. For example, \cite{NI2019358} leverages GNNs to adaptively allocate inflows, learning to prioritize areas with higher demand while imposing penalties for exceeding total inflow limits. 
\end{itemize}

Apart from the two-stage heterogeneous PC above, there are also one-stage heterogeneous PC approaches. N-MP \cite{liu2024nmp} combines MaxPressure with heterogeneous PC by considering multi-hop downstream links' queue lengths beyond perimeter intersections. When the average density of the protected region exceeds the critical density, N-MP deprioritizes inflows from those perimeter intersections having higher aggregated congestion level at multi-hop downstream links.

Our work is independently developed and differs from N-MP in several key aspects. Specifically, our approach emphasizes the mathematical formalization of multi-hop downstream pressure using Markov chain theory, which inherently accounts for the decaying influence of higher-hop traffic links. In contrast, N-MP employs an equal weighting scheme that does not incorporate this decaying effect.

\subsection{Macroscopic fundamental diagrams}
At the \textit{network level}, Macroscopic Fundamental Diagrams (MFDs) describe the relationship between traffic production and congestion levels for homogeneous urban networks. Originally proposed by \cite{godfrey1969mechanism} and reintroduced by \cite{daganzo2007urban, daganzo2008analytical, daganzo2011macroscopic}, MFDs have been validated through experimental findings, such as those in Yokohama, Japan, demonstrating their practical applicability \cite{geroliminis2008existence}. MFDs have been widely applied in studies focused on congestion pricing \cite{zheng2012dynamic} and traveler information dissemination \cite{mahmassani2013urban}. The most notable application of MFDs is in perimeter control, which has been discussed comprehensively in Section \ref{sec:lit-review-pc}.

Despite their widespread use, MFDs come with inherent limitations. The reliability of MFDs diminishes in regions with high spatial heterogeneity in congestion distribution, which adversely impacts their accuracy \cite{mazloumian2010spatial, mahmassani2013urban, geroliminis2011properties}. Additionally, MFD accuracy can deteriorate if detectors are not ideally placed within the network \cite{buisson2009exploring}. \cite{ji2010investigating} further demonstrated that MFDs in urban-highway hybrid networks can be scattered and heavily influenced by the traffic control plan.

While MFDs are beneficial for high-level strategic traffic management, their broad network-level perspective fails to capture localized traffic conditions around specific areas, such as feeder links. This lack of spatial granularity poses a challenge for applications requiring more detailed, localized control, particularly in addressing the heterogeneous congestion of complex traffic scenarios.

\subsection{Traffic pressure}
At the \textit{intersection level}, traffic pressure quantifies the traffic statistic difference (such as vehicle count or density) between upstream and downstream links
\cite{zhang2012traffic}, which is inspired by works \cite{tassiulas1990stability, neely2003dynamic, georgiadis2006resource} that tackle resources reallocation in wireless communication networks. The application of traffic pressure mainly focuses the MaxPressure (MP) control policy which determines next phase activation without considering phase sequences \cite{wongpiromsarn2012distributed, varaiya2013maxpressure, varaiya2013maxpressure_springer, zaidi2015traffic, wu2017delay}, or green time assignment with fixed phase sequences \cite{tsitsokas2023two, kouvelas2014maximum, le2015decentralized, mercader2020max, levin2020max} for the next control period, and the MP control is incorporated in perimeter control strategies \cite{tsitsokas2023two, liu2022dmp}. 
In addition, traffic pressure also informed the reward design of RL-based traffic signal control \cite{wei2019presslight}.

There are many variations of definition for 1-hop traffic pressure. The link lengths are taken into account \cite{kouvelas2014maximum} so that the pressure of a short link with queued vehicles is higher than the pressure of a longer link with the same number of queued vehicles. To prioritize some phases, phase weights are proposed by \cite{xiao2015further, xiao2015throughput} in pressure definition, along with the dynamic estimation of turning ratios and saturated flows are also proposed to be adaptive to the traffic flows. To improve fairness in waiting time, \cite{wu2017delay} incorporates traffic delay into the pressure definition, with controllable relative importance between delay and queues. To avoid the practical difficulty in queue measurement, \cite{mercader2020max} proposed using travel times instead of queues to define the pressure, and are tested in both simulated and real traffic environments. Similarly, D-MP \cite{liu2022dmp} used traffic delay to define the pressure, tested to be more effective than travel time based MP. To keep the large moving platoon seamlessly travel through an intersection, C-MP \cite{ahmed2024cmp} integrated the space mean speed into pressure definition to prioritize platoon movements.  OCC-MP \cite{ahmed2024occ} prioritizes transit and other high occupancy vehicles on the MP framework to improve passenger-based efficiency. 
PQ-MP \cite{liu2024pedmp} integrates pedestrian queues into the MP framework to address the pedestrian traffic.


However, \emph{most existing variations of traffic pressure do not extend beyond immediate downstream links}. While MFDs lack the spatial granularity to capture detailed traffic dynamics, the conventional 1-hop traffic pressure is another extreme that only considers immediate downstream links without incorporating further downstream conditions. Although N-MP \cite{liu2024nmp} represents a notable step by integrating multi-hop downstream queue lengths into heterogeneous PC, it applies a uniform weighting scheme that does not account for the decaying influence of distant links. 

In contrast, our work formalizes multi-hop downstream pressure through a mathematical framework grounded in Markov chain theory, which inherently incorporates a decaying influence from higher-hop traffic links. This novel metric ensures a more realistic representation of downstream traffic dynamics. The multi-hop downstream pressure has customizable spatial granularity to address the shortcomings of 1-hop pressure and MFDs. Informed by multi-hop pressure, we designed a simple yet efficient heterogeneous perimeter controller and tested on scenarios with different levels of heterogeneity. Furthermore, sensitivity analyses are conducted to evaluate the robustness of the proposed controller. 

\begin{figure*}[htbp]
\centering
  \includegraphics[width=0.8\textwidth]{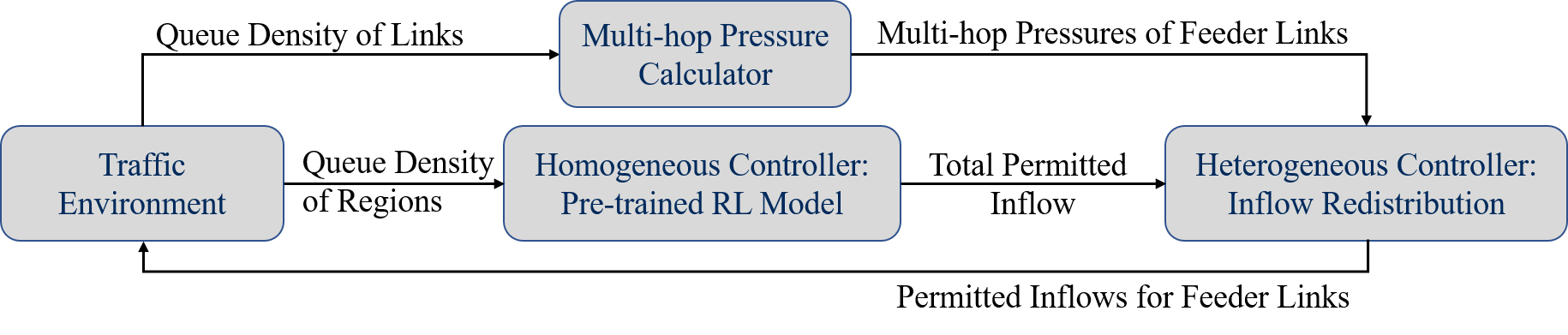}
\caption{Two-stage control scheme: The first stage is a homogeneous controller (a pre-trained deep reinforcement learning model) that generates total permitted inflow. The second stage is a heterogeneous controller that redistributes the total permitted inflow as per multi-hop pressure at each feeder link.}
\label{fig:method-scheme}
\end{figure*}



\section{Methodology}
The outline of this section is as follows: First, we summarize key mathematical notations. Then, we present our graph-based representations of traffic networks, highlighting the incorporation of a ``supersink'' for destination aggregation to achieve the graph's adjacency matrix to be a Markov chain transition matrix and reduce the size of the graph. Following it, we propose the multi-hop pressure metric for evaluating traffic conditions over broader neighborhoods, and we provide mathematical formulations of pressure computation in the scalar form for a single link and the vectorized form for all links. Finally, this novel metric informs a heterogeneous controller based on the Softmax function to redistribute the total permitted inflow.

\subsection{Mathematical notations}
The mathematical notations are primarily introduced in this section, where we establish the essential concepts about multi-hop pressure. Notations related to  experimental setups are introduced in later sections to maintain relevance and coherence for better readability.
\begin{table}[ht]
\centering
\caption{Summary of notations and their descriptions}
\begin{tabular}{c|p{6cm}}
\hline
\textbf{Notation} & \textbf{Description} \\
\hline
$a_t^i$ & Control action for metering device $i$ at timestep $t$ \\
$Q(l)$ & Queue density of link $l$ \\
$\mathcal{F}$ & Set of feeder links \\
$\mathcal{V}$ & Set of all traffic network links (vertices in graph) \\
$\mathcal{E}$ & Set of edges in graph representation \\
$\Omega$ & Supersink \\
$\mathcal{V}^e$ & Extended set of traffic links including supersink \\
$\mathcal{E}^e$ & Extended set of edges \\
$G = (\mathcal{V}, \mathcal{E})$ & Graph representation without supersink \\
$G^e = (\mathcal{V}^e, \mathcal{E}^e)$ & Graph representation with supersink \\
$\mathcal{N}(l,h)$ & Set of $h$-hop downstream links from $l$ \\
$p(l,h)$ & $h$-hop pressure for link $l$ \\
$T_{ij}$ & Turning ratio from link $i$ to $j$ \\
$\mathbf{T}$ & Weighted adjacency matrix of graph $G$ \\
$\mathbf{P}$ & \begin{tabular}[c]{@{}l@{}}Weighted adjacency matrix of graph $G^e$, which is \\  also an Markov transition matrix\end{tabular} \\
$\mathbf{Q}$ & Concatenation of queue density for links in $\mathcal{V}^e$ \\
$\mathbf{p}(h)$ & Concatenation of $h$-hop pressure \\
\hline
\end{tabular}
\label{tab:notations-for-pressure}
\end{table}

\subsection{Control architecture}
The proposed two-stage hierarchical control scheme is shown in Figure \ref{fig:method-scheme}. Similar two-stage control schemes are used in heterogeneous PC works mentioned under Section \ref{sec:lit-review-pc}.

\subsubsection{First stage: computing total permitted inflow leveraging a pre-trained RL control policy}
\begin{figure}[htbp]
\centering
  \includegraphics[width=\columnwidth]{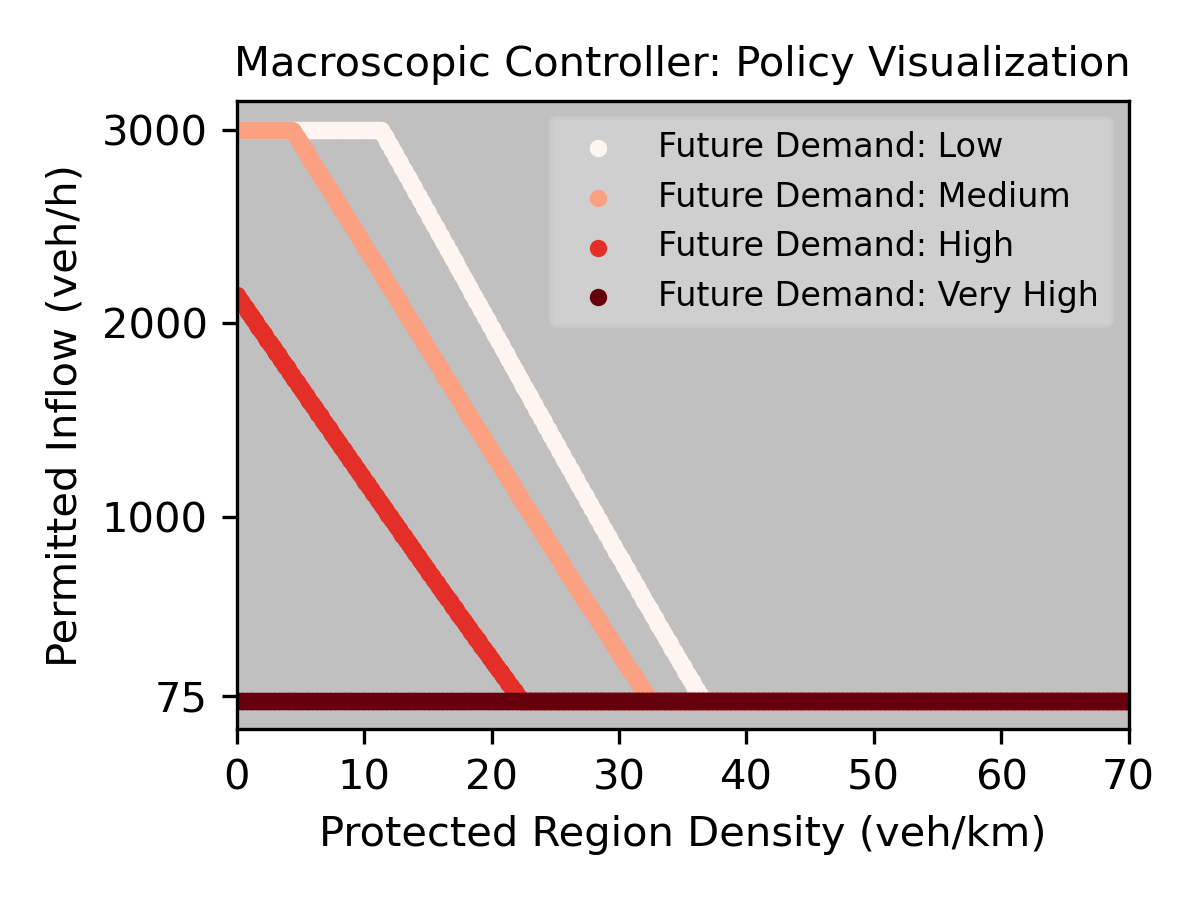}
    \caption{The pre-trained control policy for the first stage macroscopic (homogeneous) controller.}
    \label{fig:first-stage-control-policy-viz}
\end{figure}

In the first stage, a macroscopic controller computes the total permitted inflow at an aggregated level using macroscopic traffic states (e.g., traffic density of the region) from the traffic environment. This controller is a pre-trained deep reinforcement learning model \cite{li2023perimeter}, and the policy visualization is shown in Figure \ref{fig:first-stage-control-policy-viz}. In short, higher congestion in the protected region and larger future demand lead to more restricted inflow.

\subsubsection{Second stage: redistributing total permitted inflow leveraging multi-hop pressures}
In the second stage, the multi-hop pressure calculator computes multi-hop pressures of feeder links around the perimeter using microscopic traffic states (e.g., traffic density of each link). At the end of each control cycle, the total permitted inflow is redistributed by the heterogeneous controller, requiring the multi-hop pressures of feeder links. The controller must satisfy these two requirements:
\begin{itemize}
    \item \textit{Monotonous increase relationship between pressure and control action}: a feeder link with larger pressure should be assigned with a larger permitted inflow rate since a larger pressure indicates more possibly unused capacities at downstream links.
    \item \textit{Total permitted inflow conservation}: The summation of all feeder links permitted inflows should equal to the total permitted inflow.
\end{itemize}

Following the requirements, we proposed the Softmax controller for its simplicity. We define \textit{the perimeter $h$-hop pressure vector}  $\mathbf{p}_t^{\mathcal{F}} (h) \in \mathbb{R}^{|\mathcal{F}|}$ to be the concatenation of $h$-hop pressure for each feeder link at timestep $t$:
\begin{align}
    \mathbf{p}_t^{\mathcal{F}} (h) = \oplus_{f\in \mathcal{F}} \quad p_t (f, h)
\end{align}

For feeder link $f \in \mathcal{F}$, its  permitted inflow is:
\begin{align}
    a_t^f &= A_t \times \text{Softmax}(s\mathbf{p}_t^{\mathcal{F}} (h))_f  \\
          &= A_t \times \dfrac{e^{s \times p_t (f, h)}}{\sum_{f'\in \mathcal{F}} e^{s \times p_t (f', h)}}
\end{align}
where $A_t$ is the total permitted inflow at timestep $t$ from the first stage controller, and hyperparameter $s$ (a.k.a. coldness or inverse temperature) is a positive scalar to tune the sensitivity to pressure values. When $s\rightarrow0$,  all permitted inflows are equal and the Softmax controller collapses into homogeneous control. When $s$ is large, a feeder link with higher pressure is even more prioritized by greater permitted inflow. The choice of $s$ is selected through experiments. Apparently, the summation of all feeder links' permitted inflows is the total permitted inflow owing to the property of the Softmax function. 

\subsection{Graph representations of the traffic network}
We adopt a graph representation to represent the physical traffic network.

To simplify the graph representation and enable the construction of a Markovian adjacency matrix, we introduce a \textit{supersink} $\Omega$, an abstract node merging all destinations into one, greatly reducing the number of nodes in graph. The supersink has three key properties:
\begin{itemize}
    \item \textit{Zero Queue Density}: It has infinite capacity and never rejects vehicles, so $Q(\Omega) = 0$;
    \item \textit{Absorption}: Any neighbor of the supersink, regardless of hop count, is still the supersink, i.e., $\mathcal{N}(\Omega, h) = {\Omega}$ for all $h \in \mathbb{N}$;
    \item \textit{Binary Turning Ratios}: The turning ratio from the supersink to itself or from an exit link (a link solely connected to $\Omega$) to the supersink is 1, and 0 otherwise.
\end{itemize}

\begin{definition} [Graph representation]\label{def:extended-graph}
    The graph representation $G^e = (\mathcal{V}^e, \mathcal{E}^e)$, where:
    \begin{itemize}
    \item The extended link set $\mathcal{V}^e = \mathcal{V}\cup\{\Omega\}$ additionally includes a supersink vertex $\Omega$.
    \item The extended edge set $\mathcal{E}^e$ additionally includes those edges from exit links to the supersink, and the supersink to itself.
    \item Edge weight $T_{uv}$ represents the turning ratio from link $u$ to link $v$. These weights are derived from real empirical data or traffic simulations, encapsulating the probability of traffic flow transitions between links.
    \end{itemize}
\end{definition}

\begin{figure}[htbp]
\centering
  \begin{subfigure}[b]{0.42\columnwidth}
    \includegraphics[width=\linewidth]{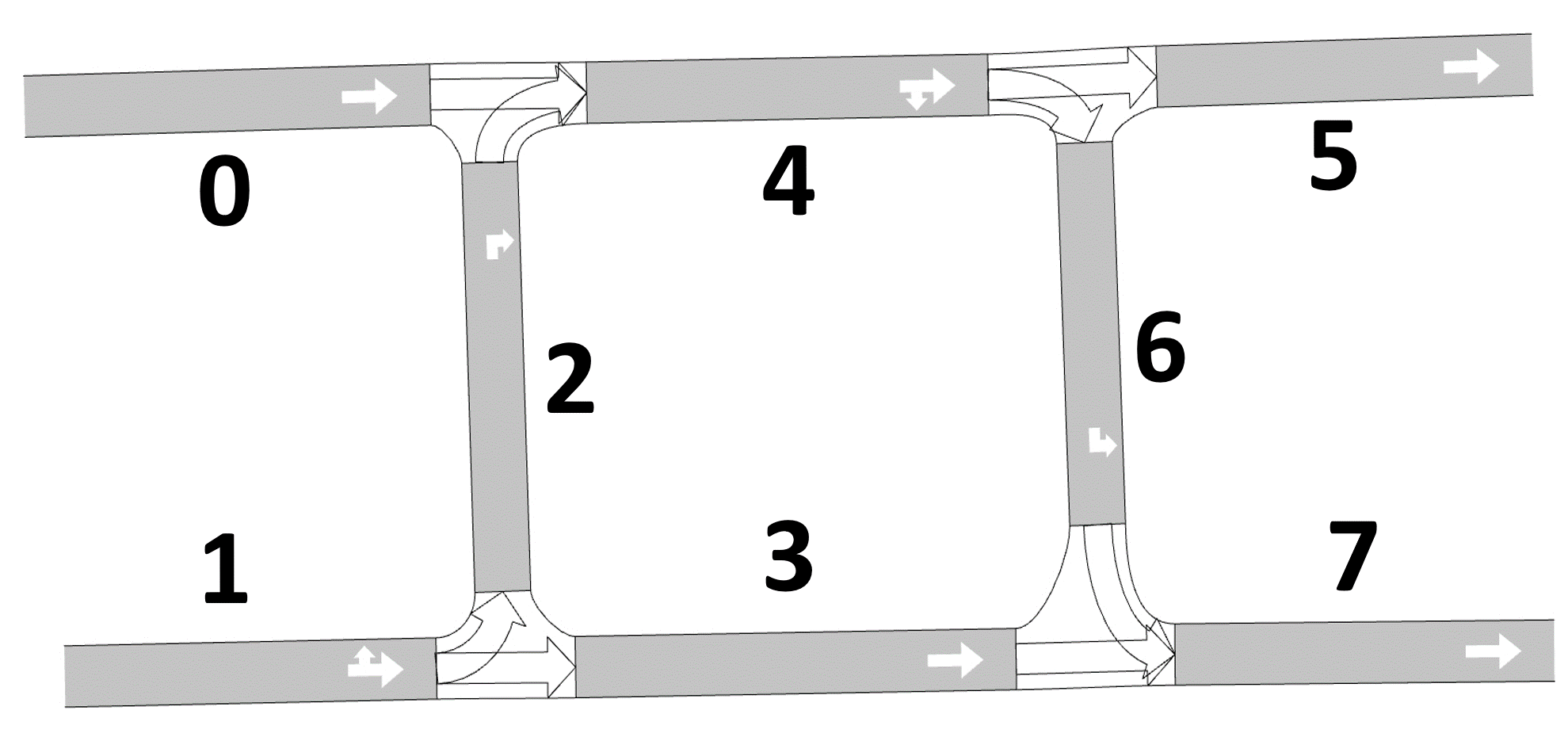}
    \caption{A toy network with 8 links.}
    \label{fig:toy-network}
  \end{subfigure}
  ~
  \begin{subfigure}[b]{0.52\columnwidth}
    \includegraphics[width=\linewidth]{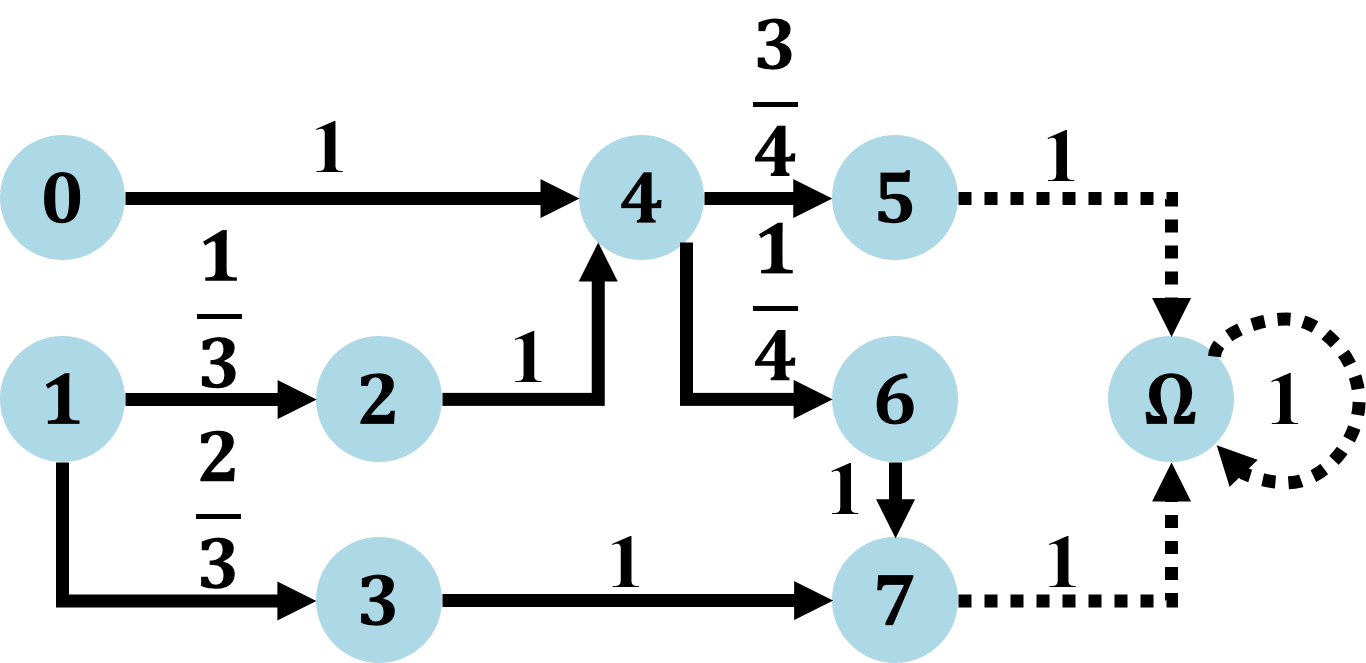}
    \caption{Graph representation of the toy network.}
    \label{fig:graph-representation-toy-network}
  \end{subfigure}
\caption{Toy network used for multi-hop pressure explanation. In Figure (a), there is no network beyond what is illustrated. In Figure (b), each vertex represents a traffic link, and each edge represents the connection of links, and the weights shown on the edges are the fabricated turning ratios. The vertex $\Omega$ is the supersink, and the edges in dashed lines represent graph $G^e$ is extended from graph $G$.}
\vspace{-5pt}
\end{figure}

To provide an example of graph representation, a simplistic traffic network with 8 links, shown in Figure \ref{fig:toy-network}, is mapped onto its graph representation depicted in Figure \ref{fig:graph-representation-toy-network}.

\subsection{Vehicle Movement as an Absorbing Markov Chain}

A vehicle traveling in a traffic network according to certain turning ratios can be formalized as a time-homogeneous absorbing Markov chain by treating a vehicle presenting on a link $l$ as a random variable with probability $Pr(x=l)$. The state space of this Markov chain is a finite set with size $|\mathcal{V}^e|$, which is the total number of traffic links in the entire network. The transition matrix $\mathbf{P}$ is exactly the weighted adjacency matrix of graph $\mathcal{G}^e$, and is defined as follows:
\begin{align}\label{eq:markov-chain-matrix}
    \mathbf{P} = \left[\begin{array}{ccc:c}
        T_{11} & \ldots & T_{1|\mathcal{V}|} & T_{1\Omega}\\
        \vdots & \ddots & \vdots & \vdots \\
        T_{|\mathcal{V}|1} & \ldots & T_{|\mathcal{V}||\mathcal{V}|} & T_{|\mathcal{V}|\Omega}\\
        \hdashline
        0      & \ldots & 0      & 1
    \end{array}\right] = 
    \begin{bmatrix}
        \mathbf{T}               & \mathbf{T}_{*\Omega} \\
        \mathbf{0}^\top & 1
    \end{bmatrix},
\end{align}
where $\mathbf{T}_{*\Omega}$ is a concatenation of binary turning ratios related to supersink.

\subsection{Multi-hop downstream pressure: a customizable metric for long-distance traffic condition}
Congestion on a nearby street has a more immediate and significant impact than congestion several blocks away.
Therefore, the multi-hop pressure definition needs to capture the diminishing influence of distant congestion while still accounting for its cumulative effect on the current traffic link. To understand the downstream links at different hops, the set of downstream links for link 0 in Figure \ref{fig:toy-network} at different hops can be written as follows:
\begin{align}
    \mathcal{N}(0, 0) &= \{0\}\\
    \mathcal{N}(0, 1) &= \{4\}\\
    \mathcal{N}(0, 2) &= \{5,6\}\\
    \mathcal{N}(0, 3) &= \{7\}\\
    \mathcal{N}(0, h) &= \{ \Omega \}, \quad h\geq 4, h\in\mathbb{N}^+
\end{align}

The scalar version of multi-hop pressure, which computes pressure for a single link, is less compact and therefore included in the Appendix. Unlike the scalar version, the vectorized version can compute pressures for \textit{all} links in the traffic network simultaneously. Therefore, the vectorized form accelerates the computation upon implementation compared to iterating over each link in the network:
\begin{tcolorbox}[colback=lightgray!20, colframe=black, sharp corners]
\begin{align}
    \mathbf{p}(0) &= \mathbf{Q} & \label{eq:markov-0-hop} \\
    \mathbf{p}(h) &= \mathbf{p}(h-1) - \mathbf{P}^h\mathbf{Q}, & h\in\mathbb{N}^+ &\quad \text{(Recursive form)} \label{eq:markov-recursive-matrix-pressure} \\
    \mathbf{p}(h) &= \mathbf{Q} - \sum_{h'=1}^{h}\mathbf{P}^{h'} \mathbf{Q}, & h\in\mathbb{N}^+ &\quad \text{(Unrolled form)} \label{eq:markov-unroll-matrix-pressure}
\end{align}
\end{tcolorbox}

When $h=1$, Eq. (\ref{eq:markov-unroll-matrix-pressure}) collapses into the vanilla version of traffic pressure:
\begin{align}
    \mathbf{p}(h) &= \mathbf{Q} - \mathbf{P} \mathbf{Q}
\end{align}

There are two properties of multi-hop downstream pressure: (1) Monotonously decreasing over the hop and (2) Bounded range, whose proof can be found in Appendix \ref{app:property-pressure}.

\noindent\textbf{Interpretations of $\mathbf{P}^h$}: It is noteworthy to articulate the term $\mathbf{P}^h$ in Eq. (\ref{eq:markov-recursive-matrix-pressure}). Generally speaking, in the Markov chain, the entry $(i,j)$ in the $h$-power of transition matrix $\mathbf{P}^h$, denoted as $(\mathbf{P}^h)_{ij}$, represents the probability of starting from vertex $i$ and reaching vertex $j$ after $h$ steps of transitions. In the context of a traffic network where entries represent turning ratios, $(\mathbf{P}^h)_{ij}$ represents the probability of a vehicle moving from link $i$ to link $j$ considering all possible paths of $h$ links:
\begin{itemize}[itemsep=4pt]
    \item $h$\textit{-hop influence}: The scalar value $(\mathbf{P}^h)_{ij}$ captures the influence that the traffic state of link $j$ has on link $i$ after $h$ transitions, that is, link $j$ is one of the $h$-hop downstream links from link $i$, i.e., $j\in \mathcal{N}(i, h)$. It embodies the idea that traffic pressure is not local but can propagate through the network from distant links. 
    
    \item \textit{Decay of influence over hop}: Since $\mathbf{P}^h$ involves higher powers of $\mathbf{P}$ as $h$ increases, the influence on pressure generally diminishes with hop due to the range of turning ratios being $[0, 1]$. This reflects the realistic attenuation of congestion effects over distance in a traffic network. 
    \item \textit{Pressure contribution}: Multiplying $(\mathbf{P}^h)_{ij}$ by $Q(j)$ incorporates the current queue density of link $j$ into the pressure calculation at link $i$. If $Q(j)$ is high (indicating congestion), and $(\mathbf{P}^h)_{ij}$ is significant, then link $j$ will contribute substantially to the pressure at link $i$. 
\end{itemize}

\section{Experimental Setup}

The proposed perimeter control scheme is evaluated using the traffic simulator AIMSUN \cite{Aimsun}. Detailed description of the experimental setup in traffic network architecture and the traffic demand are provided in this section.

\subsection{Traffic network architecture}

\begin{figure}[htbp]
\centering
  \includegraphics[width=\columnwidth]{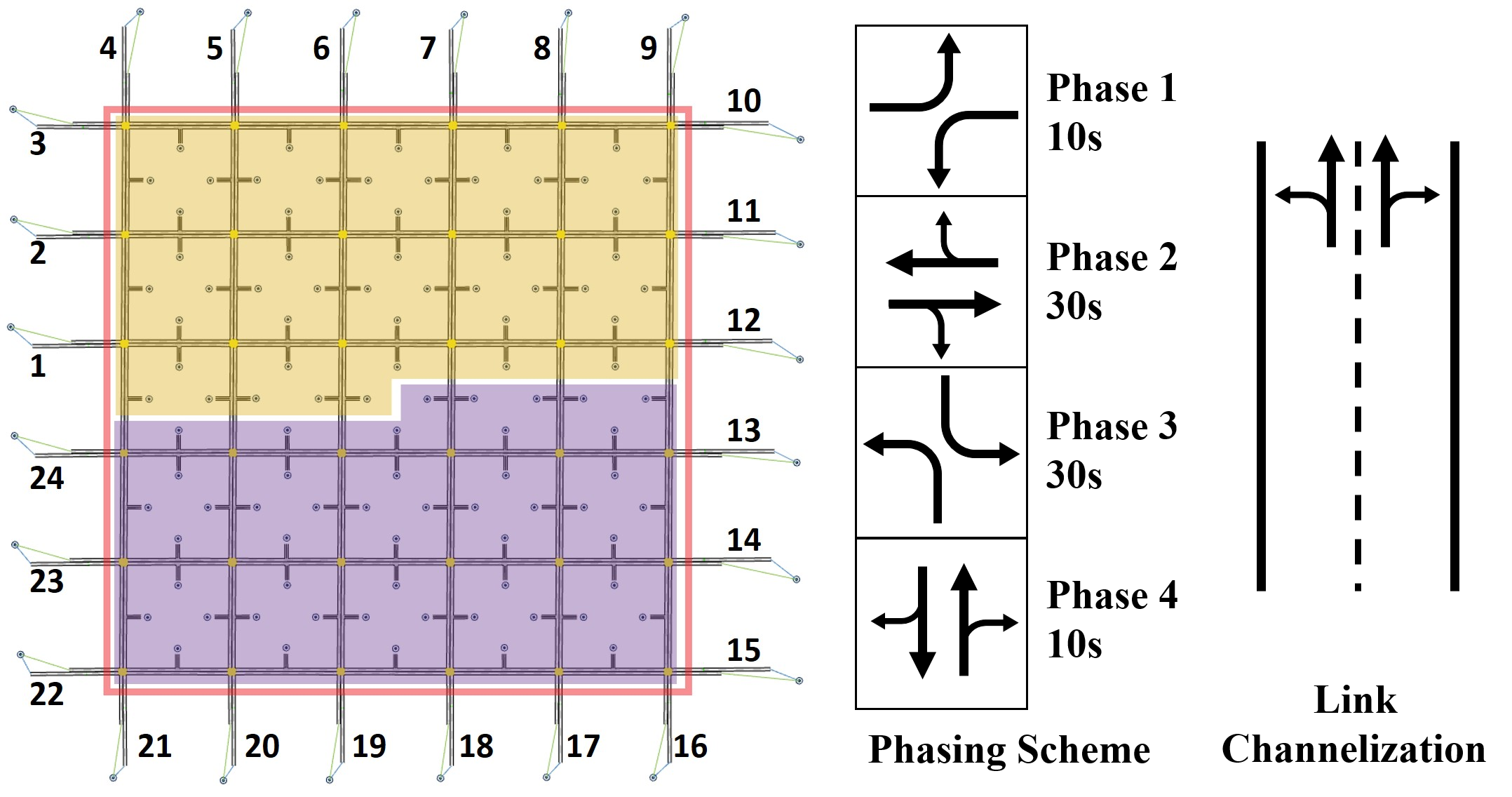}
    \caption{Tested traffic network. The protected region is bounded by a red rectangle as a perimeter. To generate imbalanced demand, the protected region is marked by two equal-sized upper and lower subregions, rendered in yellow and purple, respectively.}
    \label{fig:5x5-network}
\end{figure}

To simulate a monocentric city during peak hours, we created a grid network with 36 signalized intersections that use a fixed control plan, where the signal timings and movements in each phase are provided in Figure \ref{fig:5x5-network} along with link channelization. Similar grid networks are used in \cite{NI2019358, liu2024nmp}. In addition, there are 4 interphases with 4 seconds duration each. Hence, the cycle length is $10+30+30+10+4\times4=96$ seconds. To be consistent, the perimeter control action update cycle is 96 seconds as well. There are 24 feeder links around the perimeter indexed by integers. Each traffic link is 85 meters long (each block is $2\times85=170$ meters long) with two lanes per link and the link channelization is identical for all links, except those single-lane links (mimicking parking lot ramps) in the protected region connected to origins or destinations. Inflow metering is used for feeder links as it provides precise control over inflow without disrupting the internal signal plan.

\subsection{Heterogeneous demand setup}
Does the performance improvement relative to homogeneous control increase as the demand heterogeneity increases? To answer this research question (indexed as RQ3 in later sections), we design scenarios with different levels of demand heterogeneity. Our experiment examines two distinct approaches to modulating demand heterogeneity:
\begin{enumerate}
    \item \textit{Asynchrony} $\tau$: the time difference between upper and lower subregions' demands. As asynchrony increases, we expect a rise in congestion heterogeneity.
    \item \textit{Imbalance} $\alpha^{\text{upper}}_{22}$: the difference of demand volume generated in upper and lower subregions. As the imbalance increases, we expect a rise in congestion heterogeneity.
\end{enumerate}

\begin{figure}[htbp]
\centering
  \begin{subfigure}[b]{0.48\columnwidth}
    \includegraphics[width=\linewidth]{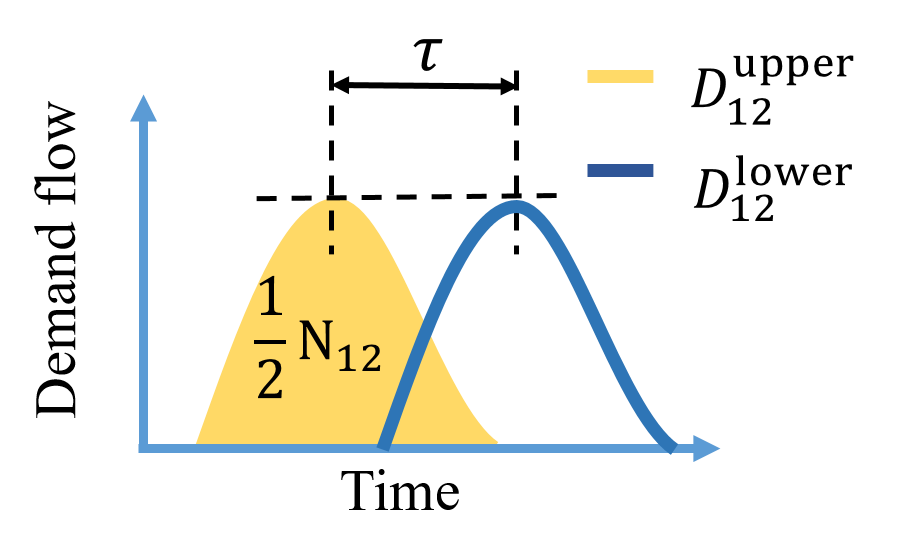}
    \caption{Profile of external demand. The external demand for upper and lower subregions are same in volume, but different in time.}
    \label{fig:external-demand}
  \end{subfigure}
  ~
  \begin{subfigure}[b]{0.48\columnwidth}
    \includegraphics[width=\linewidth]{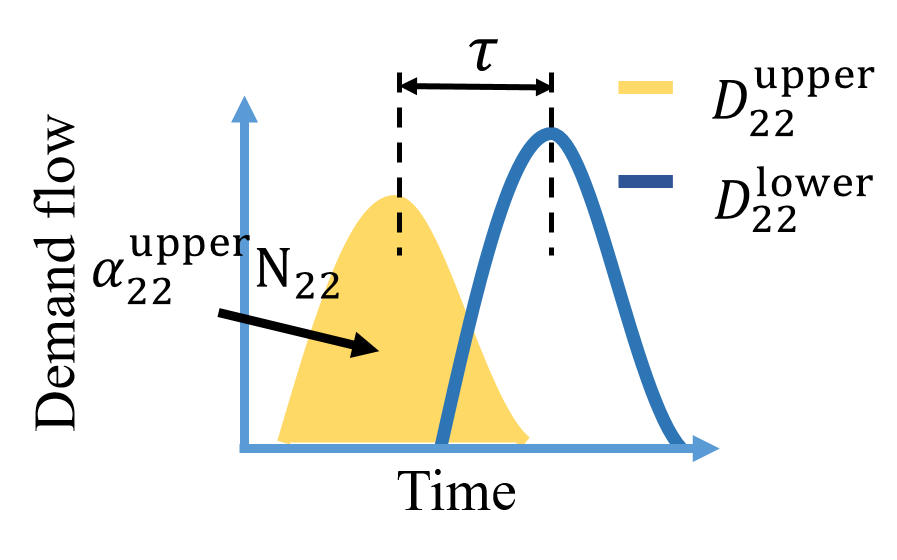}
    \caption{Profile of internal demand. The internal demand for upper and lower subregions can be different in volume and time.}
    \label{fig:internal-demand}
  \end{subfigure}
\caption{Internal and external demand profiles.}
\vspace{-5pt}
\end{figure}

To formalize the demand setup, we define mathematical notations in Table \ref{tab:notations-for-demand}. The mathematical formulation of demand is as follows:

\begin{table}[ht]
\renewcommand{\arraystretch}{1.2} 
\centering
\caption{Summary of notations for traffic demand}
\begin{tabular}{c|p{6cm}}
\hline
\textbf{Notation} & \textbf{Description} \\ \hline
$T_S$ & Total traffic simulation duration \\
$N_{12}$ & Total number of external trips \\
$N_{22}$ & Total number of internal trips \\
$D^{\text{upper}}_{22} (t)$ & Upper subregion internal demand at time $t$ \\
$D^{\text{lower}}_{22}(t)$ & Lower subregion internal demand at time $t$ \\
$\alpha^{\text{upper}}_{22}$ & Upper subregion internal demand percentage \\
$\alpha^{\text{lower}}_{22}$ & Lower subregion internal demand percentage \\
$D^{\text{upper}}_{12} (t)$ & Demand from upper feeder links to  upper subregion at time $t$ \\
$D^{\text{lower}}_{12}(t)$ & Demand from lower feeder links to  lower subregion at time $t$ \\
$\tau$ & Asynchrony parameter (time shift) \\ \hline
\end{tabular}
\label{tab:notations-for-demand}
\end{table}

\begin{align}
\int_0^{T_S} \left( D^{\text{lower}}_{12}(t) + D^{\text{upper}}_{12}(t) \right) \text{d}t &= N_{12} \label{eq:N12}
\\
\int_0^{T_S} \left( D^{\text{lower}}_{22}(t) + D^{\text{upper}}_{22}(t) \right) \text{d}t &= N_{22} \label{eq:N22} 
\\
D^{\text{lower}}_{12}(t + \tau) &= D^{\text{upper}}_{12}(t), &\forall t\label{eq:q12_async} 
\\
\dfrac{D^{\text{lower}}_{22}(t + \tau)}{\alpha^{\text{lower}}_{22}} &= \dfrac{D^{\text{upper}}_{22}(t)}{\alpha^{\text{upper}}_{22}}, &\forall t\label{eq:q22_imbalance_and_async} 
\\
\alpha^{\text{upper}}_{22} + \alpha^{\text{lower}}_{22} &= 1 \label{eq:q22_conservation} 
\\
0 < \alpha^{\text{upper}}_{22}, \alpha^{\text{lower}}_{22} &< 1 \label{eq:q22_ratio} 
\end{align}

Eq. (\ref{eq:N12}) and Eq. (\ref{eq:N22}) indicate the total number of trips for external and internal demands. Eq. (\ref{eq:q12_async}) describes the asynchrony between demands from upper and lower feeder links, as illustrated in Figure \ref{fig:external-demand}.
Eq. (\ref{eq:q22_imbalance_and_async}) describes the asynchrony and imbalance of internal demand, as illustrated in Figure \ref{fig:internal-demand}. The asynchrony is regulated by the parameter $\tau$, where a greater $\tau$ indicates later internal and external demands at the lower part of the traffic network.
For imbalance, we adjust the parameter $\alpha^{\text{upper}}_{22}$ to vary the percentage of internal demand assigned to the upper subregion. For example, $\alpha^{\text{upper}}_{22}=50\%$ means equal volume of upper and lower subregion demands. As $|\alpha^{\text{upper}}_{22}-50\%|$ increases, we expect a rise in congestion heterogeneity. Eq. (\ref{eq:q22_conservation}) and Eq. (\ref{eq:q22_ratio}) ensure the range for parameters related to imbalance. For simplification, there are no trips from the protected region to feeder links, or trips from feeder links to feeder links.

\paragraph*{Tested demands}
The tested demands are designed to address the following research questions (RQs):
\begin{itemize}
\item RQ1: How does the pressure sensitivity $s$ impact the performance of the heterogeneous controller? 
\item RQ2: What is the optimal range of hops for the best performance of the heterogeneous controller? 
\item RQ3: Does the performance improvement increase as the demand heterogeneity rises for heterogeneous control? 
\end{itemize}

For RQ1 and RQ2, the demand is set with $\tau = \frac{3}{4}$ hour and $\alpha^{\text{upper}}_{22} = 50\%$. Hence the source of demand heterogeneity comes from asynchrony, not from imbalance. The sensitivity parameter $s$ is tested with values ranging from $2^0$ to $2^7$, while the number of hops is varied from 0 to 22 in increments of 2 (each block in the traffic network requires 2 links to cross). Table \ref{tab:demand-setup} shows the enumeration of parameter $s$ and $h$. This results in $8 \times 12 = 96$ combinations of $s$ and hop.

For RQ3, the optimal controller, determined by the best combination of $s$ and hop based on minimal total time spent (TTS), is tested across various values of $\tau$ and $\alpha^{\text{upper}}_{22}$ that determines the demand heterogeneity. The $\tau$ values are $0, \frac{1}{4}, \frac{2}{4}, \frac{3}{4}, 1$, and the $\alpha^{\text{upper}}_{22}$ values range from 50\% to 80\% in 5\% increments, enumerated in Table \ref{tab:demand-setup}. This results in $5 \times 7 = 35$ combinations of $\tau$ and $\alpha^{\text{upper}}_{22}$.

Both internal and external demands are divided into nine 15-minute intervals, and the numbers of trips for these nine intervals are determined by a vector of nine weights: $[r^0, r^1, r^2, r^3, r^4, r^3, r^2, r^1, r^0]$, where $r$ (fixed to 2) serves as a hyperparameter to regulate the peakedness of the demand profile. The percentage for each interval can be readily obtained by normalizing the vector by dividing the total weights. The total external and internal trip counts are set as $N_{12} = 6000$ (veh) and $N_{22} = 11000$ (veh), respectively. These trips counts are consistent among all tested scenarios.

\begin{table}[ht]
\renewcommand{\arraystretch}{1.2} 
\centering
\caption{Summary of parameters for tested demands} 
\begin{tabular}{c|c} 
\hline \textbf{Parameter} & \textbf{Tested Values} \\ \hline Sensitivity $s$ & $2^0, 2^1, 2^2, 2^3, 2^4, 2^5, 2^6, 2^7$ \\ 
Number of hops $h$ & $0, 2, 4, \dots, 22$ (increments of 2) \\ 
Asynchrony $\tau$ & $0, \frac{1}{4}, \frac{2}{4}, \frac{3}{4}, 1$ (hour) \\ 
Imbalance $\alpha^{\text{upper}}_{22}$ & 50\%, 55\%, 60\%, 65\%, 70\%, 75\%, 80\% \\ \hline
\end{tabular} 
\label{tab:demand-setup}
\end{table}

\subsection{Compared baselines}
\begin{enumerate}
    \item \textit{Homogeneous PC} \cite{li2023perimeter}: This pre-trained RL-based control policy shown in Figure \ref{fig:first-stage-control-policy-viz} operates more smoothly than the Bang-bang control that suffers from oscillation, while being more proactive and adaptive than the reactive PI feedback control.
    \item \textit{Heterogeneous PC with Myopic Pressure}:  As a variant of our proposed method, this baseline represents a heterogeneous PC approach where the total permitted inflow is redistributed based on the vanilla myopic pressure. It serves as a straightforward comparison point to highlight the benefits of extending to multi-hop pressure.
    \item \textit{Clustered N-MP} \cite{liu2024nmp}: This baseline employs an equal weighting scheme on multi-hop downstream links (cluster) using simple average when the cluster's queue density exceeds a critical density. Hence, unlike our approach, neither influence decay over high hops nor turning ratios are integrated in N-MP. To align the experimental setup, we implement the Clustered N-MP style weighting scheme based on queue density. To the best of our tuning ability, we aimed to ensure that N-MP performed optimally as a comparative baseline against our method. 
    
\end{enumerate}

\section{Results \& discussion}
\begin{figure}[htbp]
\centering
  \includegraphics[width=0.8\columnwidth]{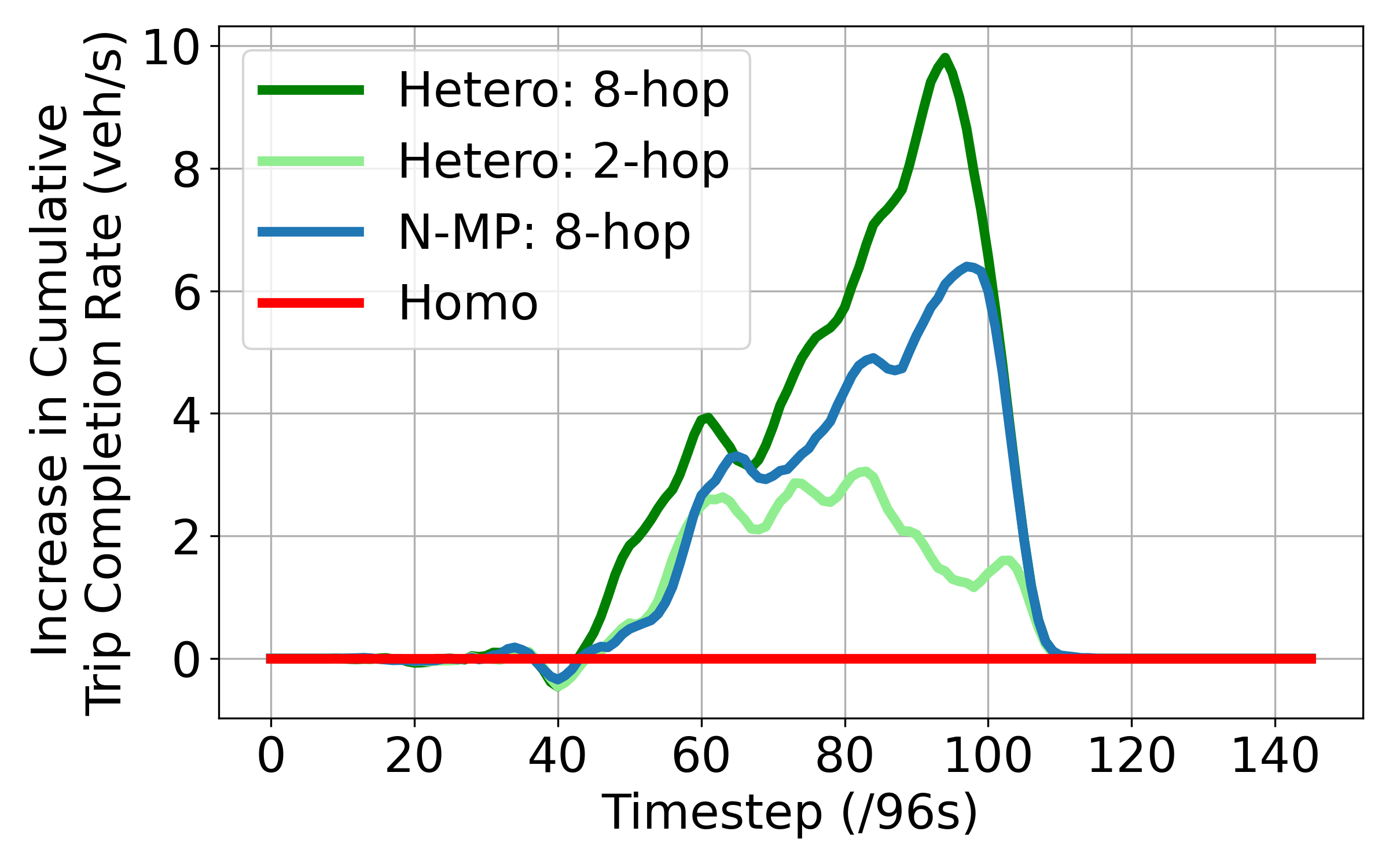}
    \caption{Increase in cumulative trip completion rate relative to homogeneous control.}
    \label{fig:tcr-increase}
\end{figure}

\begin{figure}[htbp]
\centering
  \includegraphics[width=0.8\columnwidth]{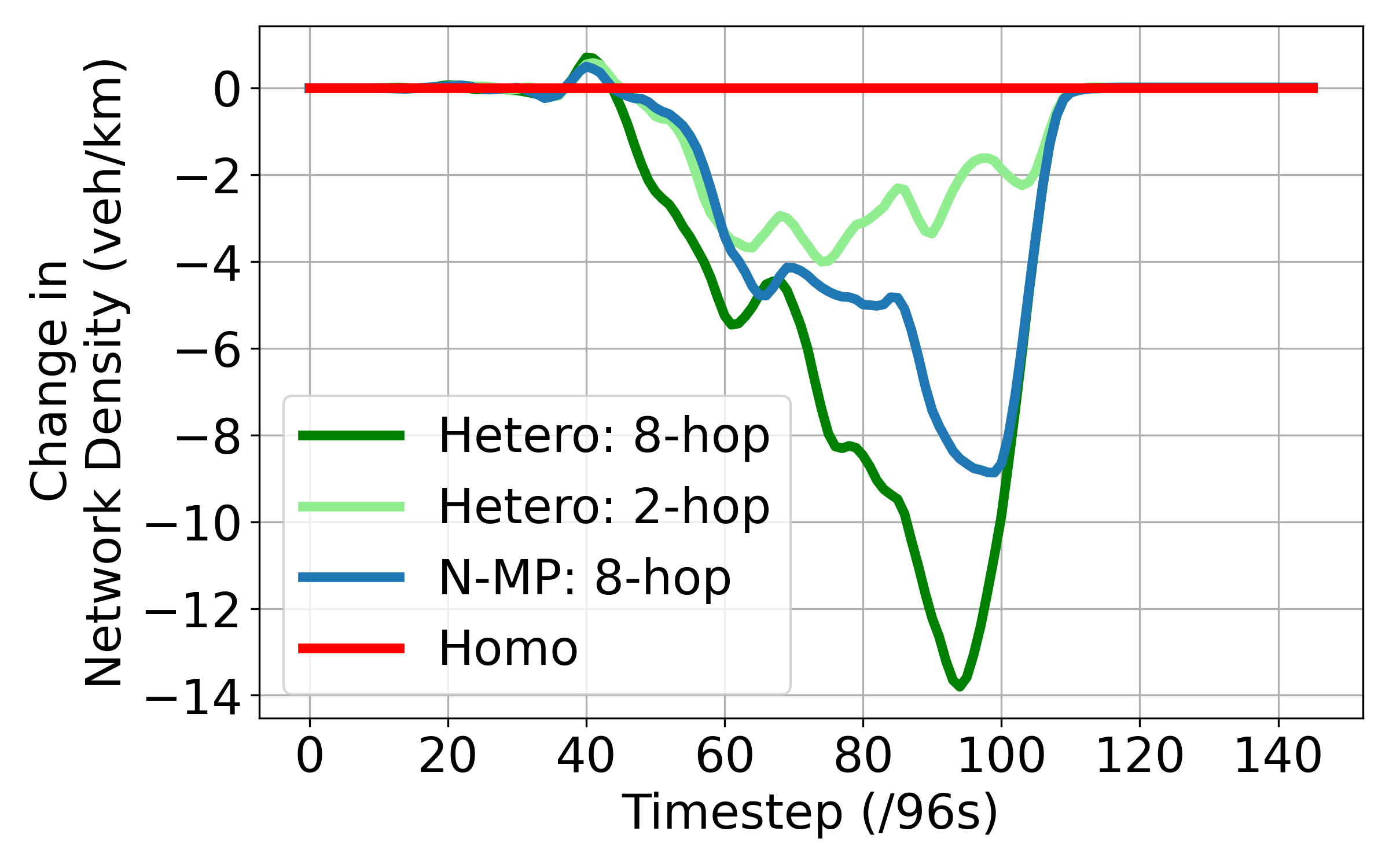}
    \caption{Change in network density relative to homogeneous control.}
    \label{fig:density-decrease}
\end{figure}

\paragraph*{Comparison with baselines} The results demonstrate that our 8-hop approach outperforms both the myopic version (2-hop) and the N-MP baseline across multiple metrics. Table \ref{tab:tts_comparison} shows that our 8-hop method achieves the lowest total time spent (TTS), with significant improvements in the protected region, while slightly higher TTS outside compared to three baselines. Figure \ref{fig:tcr-increase} and Figure \ref{fig:density-decrease} further highlight the superiority of our 8-hop approach in handling network-wide traffic. Our 8-hop strategy achieves the highest cumulative trip completion rate (Figure \ref{fig:tcr-increase}) and the most substantial reduction in network density (Figure \ref{fig:density-decrease}) relative to homogeneous control.

The performance of our 8-hop approach attributes to the incorporation of a broader spatial horizon, capturing the impact of multi-hop downstream traffic conditions more effectively. Unlike the myopic 2-hop method, which considers a limited spatial scope, and N-MP, which applies an equal weighting scheme without accounting for the decaying influence of higher hops, our 8-hop approach integrates multi-hop pressures with a mathematically grounded Markov chain framework. This allows a more informed redistribution of inflows.

\begin{table}[h]
\centering
\caption{Comparison of TTS (hour)}
\begin{tabular}{c|c|c|c|c}
\hline
\textbf{Method} & \begin{tabular}[c]{@{}l@{}}\textbf{Hetero:} \\\textbf{8-hop} \end{tabular} & \begin{tabular}[c]{@{}l@{}}\textbf{Hetero:} \\\textbf{2-hop} \end{tabular} & \begin{tabular}[c]{@{}l@{}}\textbf{N-MP:} \\\textbf{8-hop} \end{tabular} & \textbf{Homo} \\ \hline
\textbf{TTS} & \textbf{2457}                   & 2933                   & 2686                 & 3226          \\ \hline
\textbf{TTS Inside} & 1979                   & 2643                   & 2447                 & 2883          \\ \hline
\textbf{TTS Outside} & 478                   & 290                   & 239                 & 343          \\ \hline
\end{tabular}
\label{tab:tts_comparison}
\end{table}

\begin{figure}[htbp]
\centering
  \includegraphics[width=0.5\textwidth]{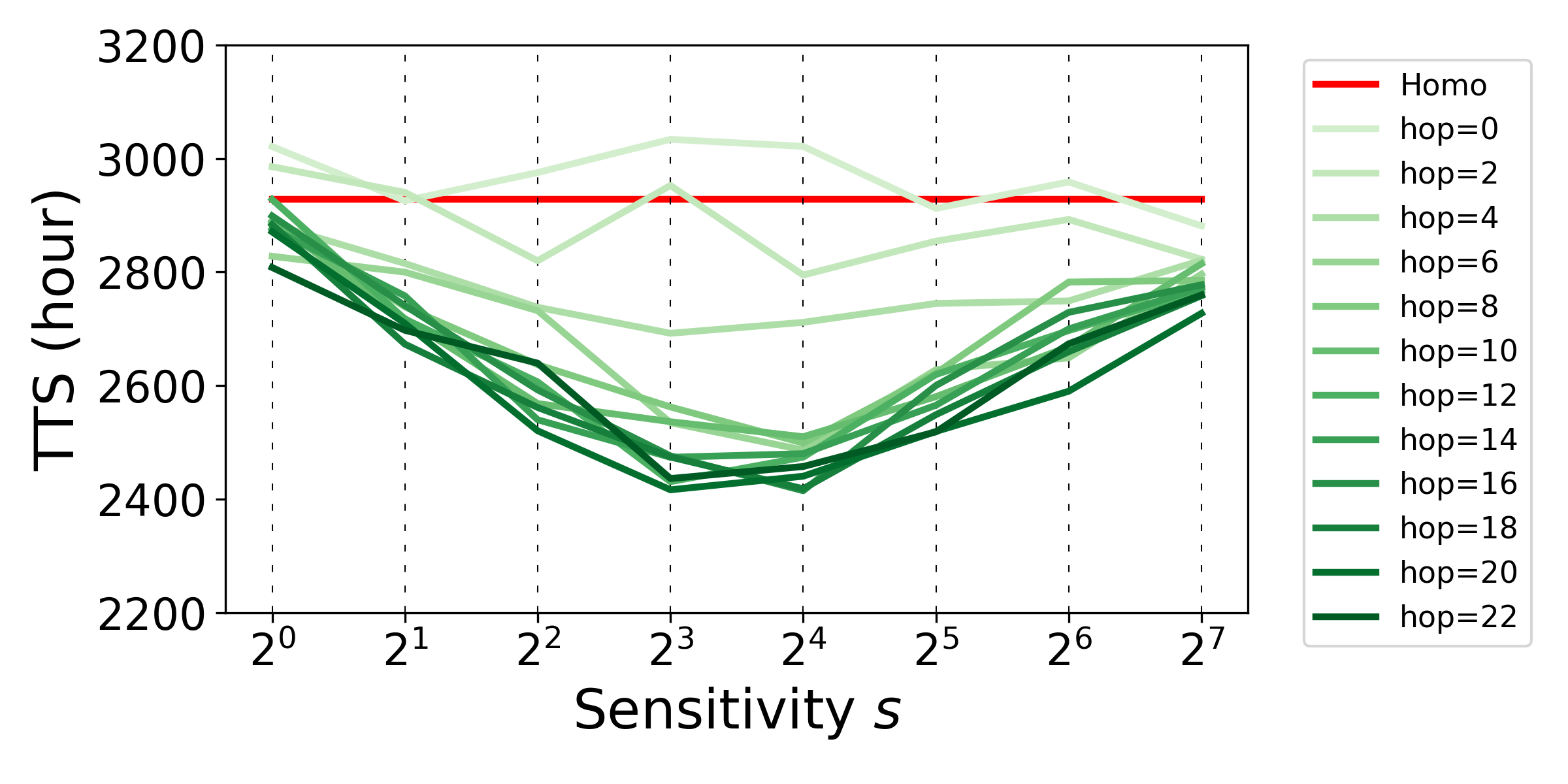}
    \caption{TTS vs. sensitivity for the heterogeneous controller.}
    \label{fig:softmax-tts-vs-sensitivity-all-hops}
\end{figure}

\paragraph*{RQ1: Calibration of control parameter $s$} Figure \ref{fig:softmax-tts-vs-sensitivity-all-hops} shows the TTS for the heterogeneous controller across different values of pressure sensitivity $s$. The results, averaged over simulations with 10 random seeds to account for traffic stochasticity, indicate that the performance improves significantly when $s$ is within an optimal range between $2^3$ and $2^4$. Outside this range, performance drops, as both overly low and excessively high values of $s$ negatively impact the controller's effectiveness. This highlights the importance of calibrating the control parameters to optimize performance.

\paragraph*{RQ2: Optimal hops and performance convergence} The TTS results in Figure \ref{fig:softmax-tts-vs-sensitivity-all-hops} also reveal that performance improves as the number of hops increases, showing a trend of gradual enhancement compared to the baseline of homogeneous control (labeled 'Homo'). The performance converges after approximately 10 hops, indicating that beyond this range, there is minimal additional benefit. 

\begin{figure}[htbp]
\centering
  \includegraphics[width=0.8\columnwidth]{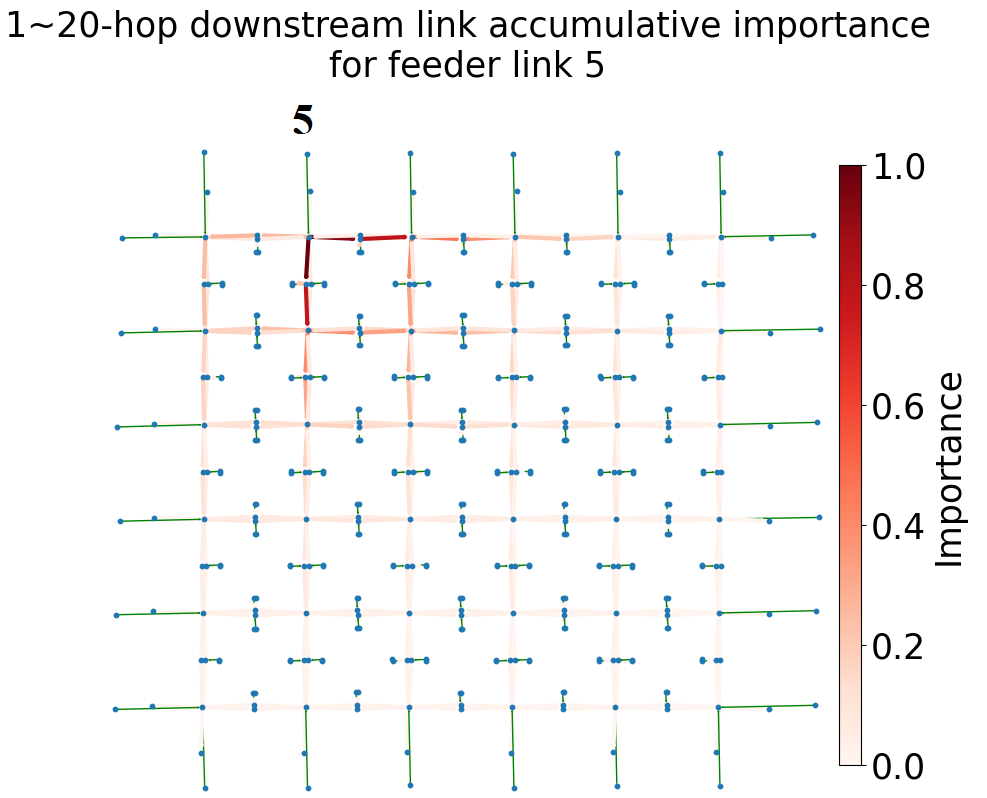}
\caption{Heatmap of 20-hop downstream link accumulative importance for feeder link 5. The importance decays over hops, implying those near downstream links have a greater impact than those far downstream links.}
\label{fig:decay-impact-5x5-network}
\end{figure}

\paragraph*{Decayed impact of downstream links over hops} The performance convergence can be further explained by the decaying influence over hops. Figure \ref{fig:decay-impact-5x5-network} visualizes the 20-hop downstream link accumulative importance for feeder link 5. The accumulative importance reflects how traffic conditions from 1-hop to 20-hop downstream links impact the upstream traffic at feeder link 5. A darker color of a downstream link indicates a greater impact on feeder link 5. A distinct decay in importance with increasing hop is observed, emphasizing a trend where proximal downstream links exert a more pronounced effect on the feeder link than those distant downstream links. The decay pattern provides insight into the performance convergence beyond 10 hops: as the importance of downstream links diminishes over hop, the incremental contribution beyond 10 hops becomes increasingly negligible. This analysis validates the importance of downstream links in proximity and the rationale behind employing a finite spatial scope in multi-hop pressure.

\begin{figure}[!htbp]
\centering
\includegraphics[width=\columnwidth]{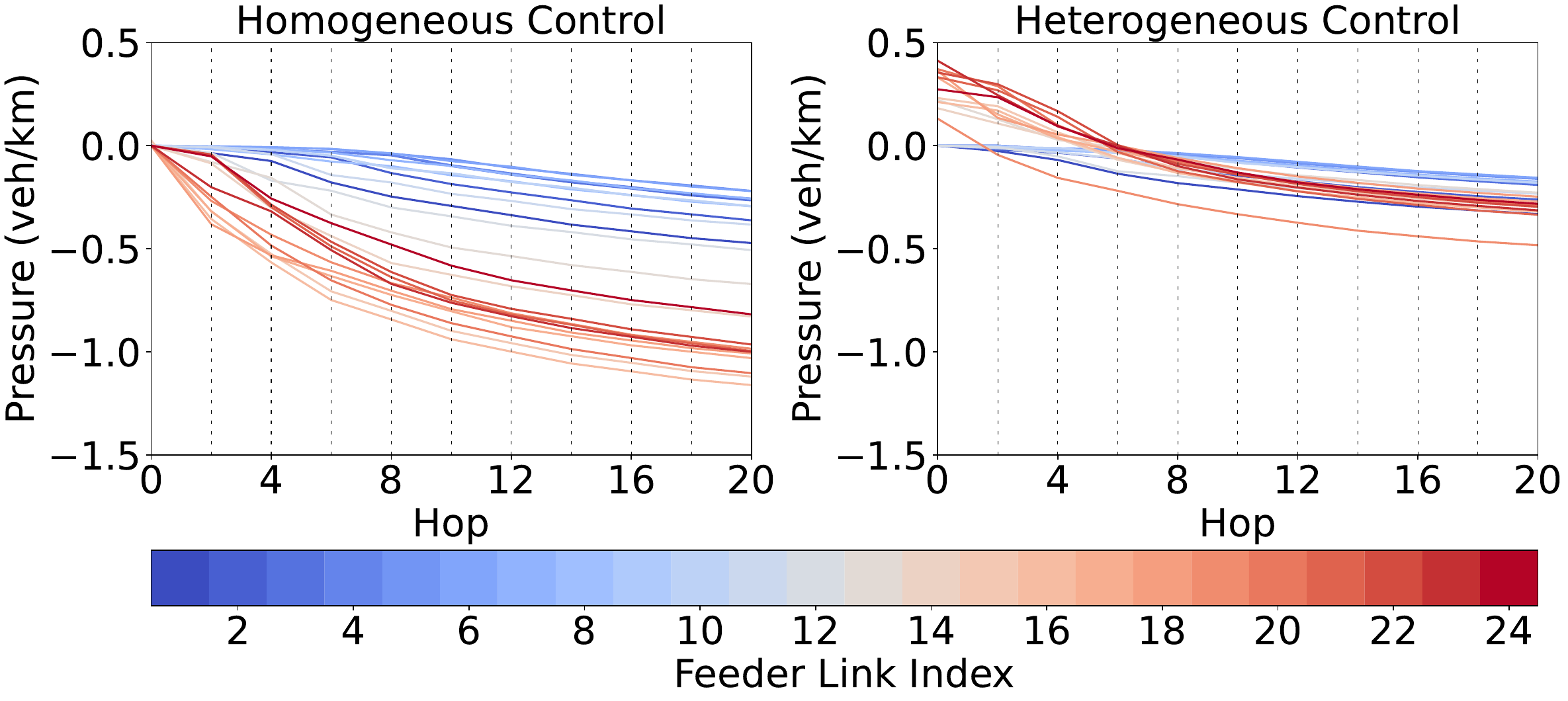}
    \caption{Profile of pressures for feeder links under homogeneous control and heterogeneous control. The heterogeneous control has more balanced pressure profiles than the homogeneous control; the heterogeneous control has smaller pressure profiles (in absolute value) than the homogeneous control.}
\label{fig:profile-pressure-homo-and-softmax-control}
\end{figure}

\paragraph*{Profile of pressures for each feeder link} In Figure \ref{fig:profile-pressure-homo-and-softmax-control}, we compare the traffic pressure profiles under homogeneous control and heterogeneous control. Both left and right figures capture the pressure at different hops for each feeder link at the same timestep. Under homogeneous control, the pressure profile on the left figure indicates a relatively imbalanced distribution across the feeder links, especially upper feeder links labeled in blue and lower feeder links labeled in red. This is because the homogeneous control cannot differentiate among the feeder links' varying congestion levels. The negative pressure with a large magnitude for lower feeder links indicates very congested downstream conditions. 
Conversely, the heterogeneous control demonstrates an equalized pressure distribution across feeder links. The small magnitude of pressures indicates the feeder link and its downstream have similar traffic potentials, and the downstream is not too congested. The heterogeneous controller modulates the total permitted inflow more adaptively by allocating higher permitted inflow to a feeder link with larger pressure (possibly an uncongested downstream) and reducing the permitted inflow to a feeder link with smaller pressure (possibly a congested downstream), potentially alleviating congestion in both feeder links and downstream.


\begin{figure}[htbp]
\centering
  \includegraphics[width=0.95\columnwidth]{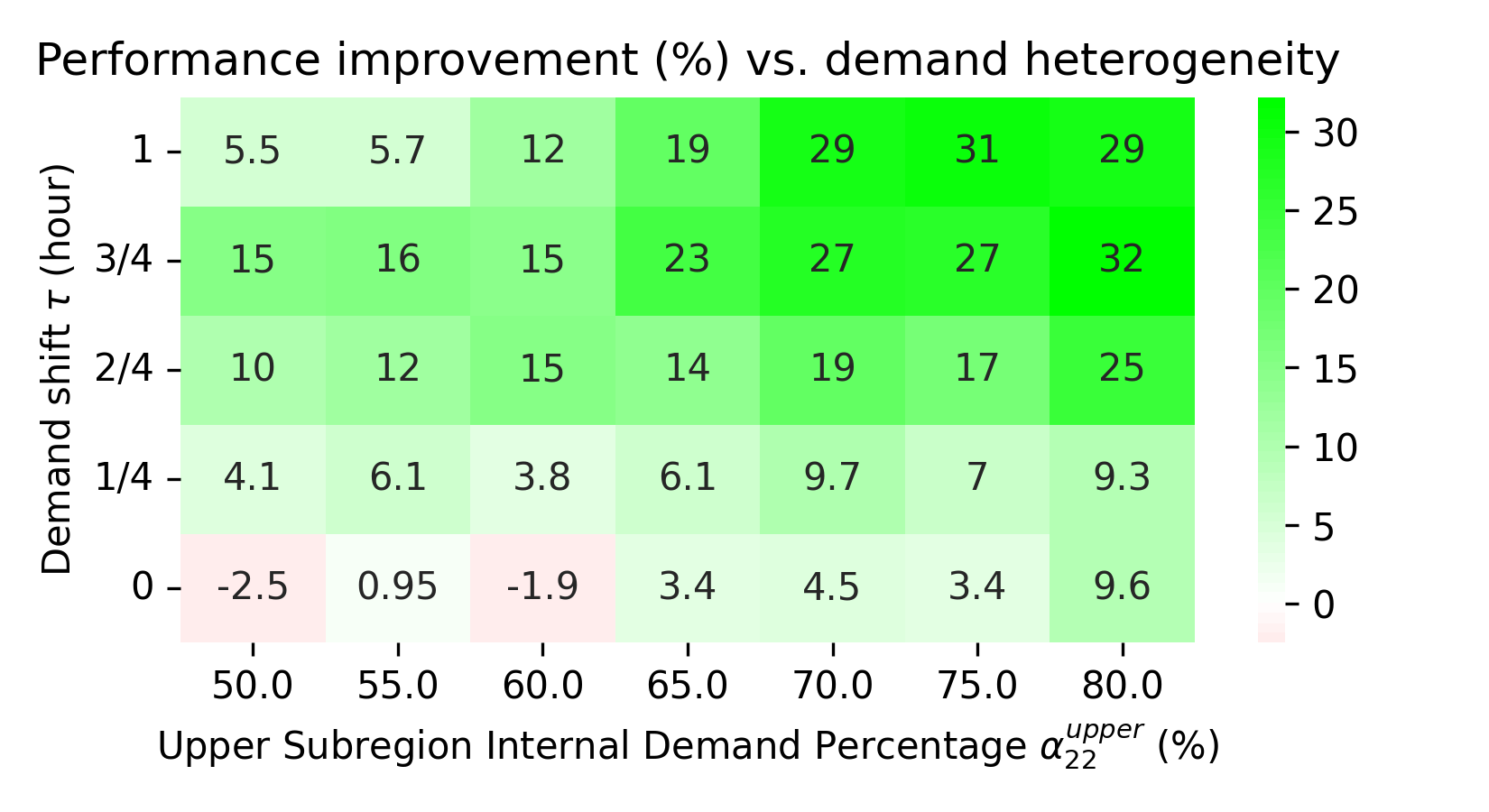}
    \caption{Performance improvement of the heterogeneous controller relative to the homogeneous control. Numbers in cells represent the performance improvement in percentage. The top right corner represents the most heterogeneous demand setup, while the bottom left corner represents the most homogeneous ones. The greater the demand heterogeneity, the better the performance of the heterogeneous controller.}
    \label{fig:tts-vs-hetero-softmax-heatmap}
\end{figure}

\paragraph*{RQ3: Performance vs demand heterogeneity} Figure \ref{fig:tts-vs-hetero-softmax-heatmap} shows a heatmap of how the performance improvement relates to two sources of demand heterogeneity: the demand shift $\tau$ (asynchrony) and the upper subregion internal demand percentage $\alpha^{\text{upper}}_{22}$ (imbalance). The demand setup on the top right is the most heterogeneous, while the bottom left is the most homogeneous. The gradient of colors, ranging from green to red, conveys the magnitude of improvement, with more intense greens indicating higher performance improvement. Generally speaking, an increase in demand heterogeneity correlates with a more pronounced improvement in performance. Conversely, demand setup with lower heterogeneity exhibits minimal or slightly negative performance improvement denoted by the light green and red cells. This trend demonstrates that the proposed heterogeneous controller is more effective in a more heterogeneous scenario. 


\section{Sensitivity analysis: robustness against turning ratio perturbations}

\paragraph*{RQ4: Robustness against turning ratio perturbation: Is the heterogeneous PC resilient to uncertainties in turning ratio estimations?} The objective of this section is to assess the robustness of the multi-hop pressure-based approach when faced with turning ratio perturbations.

\paragraph*{Experimental setup}
The experimental design involves sampling turning ratios for each link in the traffic network. For each traffic simulation, the turning ratio of each link is sampled according to a probabilistic distribution and fixed during that simulation. The simulation is repeated 20 times for each distribution at a specified perturbation level to gather statistical data. To ensure the only random variable is the turning ratio, the simulation seed is fixed to 1.

\paragraph*{Turning ratio perturbations} The perturbed turning ratios are sampled from a Dirichlet distribution (See Eq. (\ref{eq:dirichlet})) in which its sample naturally satisfies the property of probability space. The mean of the distribution is parameterized by a weighted average of the original turning ratios $\mathbf{T}_i = \oplus_{j\in \mathcal{N}(i, 1)} \quad T_{ij}$ and a uniform vector $\mathbf{1}_i = \oplus_{j\in \mathcal{N}(i, 1)} \quad 1$. The parameter $M=100$ is the inverse temperature that regulates the variance of distribution.
    \begin{align}
        \mathbf{\Tilde{T}}_i &\sim \text{Dir}(M (1-\alpha) \mathbf{T}_i + M \alpha  \mathbf{1}_i)
    \end{align}\label{eq:dirichlet}
The degree of perturbation is controlled by the parameter $\alpha \in [0,1]$, where $\alpha \rightarrow 0$ implies minimal perturbation and $\alpha \rightarrow 1$ signifies maximal perturbation. Notably, a setting of $\alpha=1$ results in the expectation of equal turning ratios, e.g., $\left[\frac{1}{3}, \frac{1}{3}, \frac{1}{3}\right]$ for a link with three turning movements. To clarify, minimal perturbation ($\alpha = 0$) does not necessarily signify no perturbation, but still results in a probabilistic distribution with non-zero entropy, where the mean of the distribution is the original turning ratio.

\begin{figure}[htbp]
\centering
  \includegraphics[width=0.7\columnwidth]{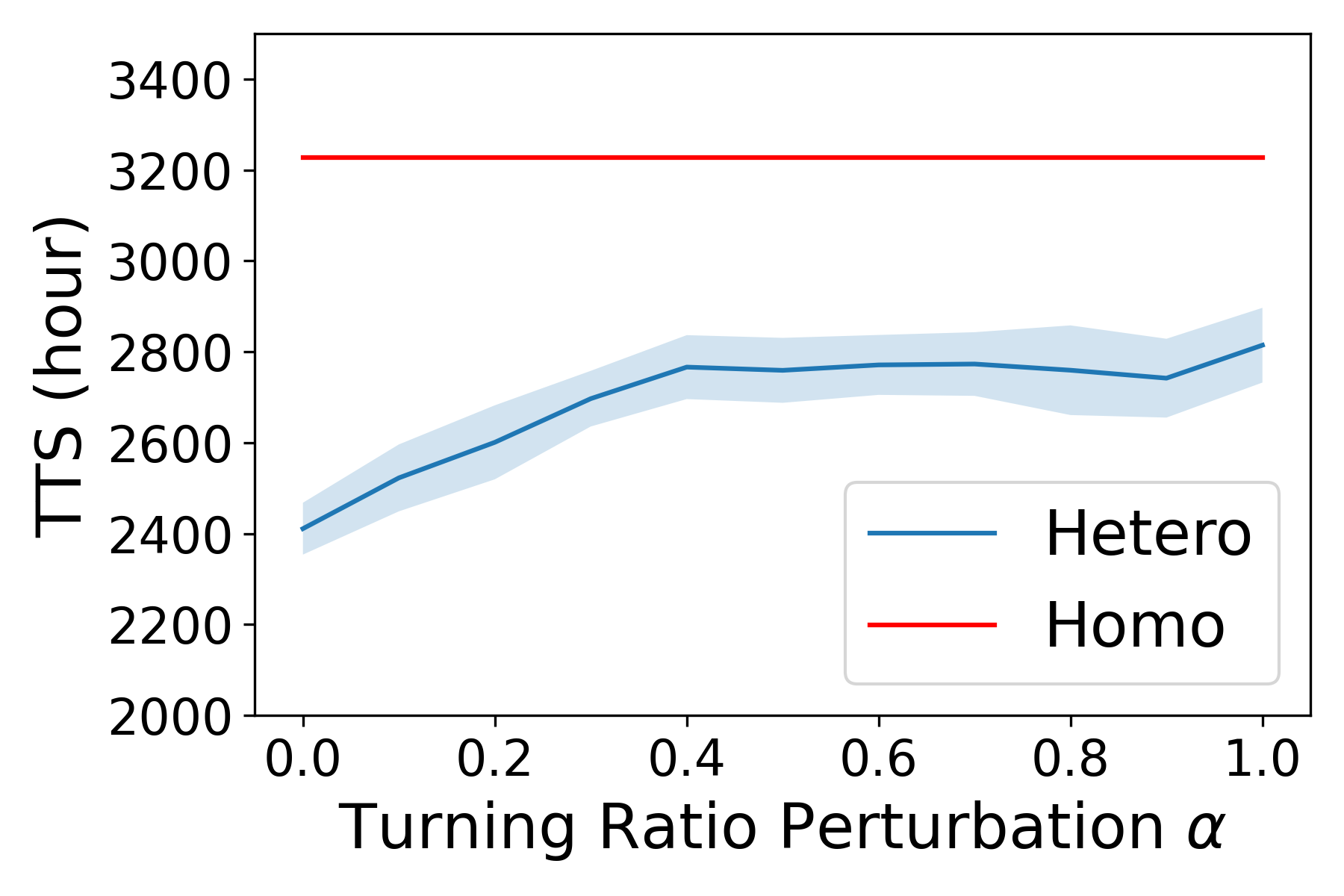}
    \caption{Robustness against turning ratio perturbations. Higher degree of perturbation leads to degraded performance, yet still outperforms homogeneous control.}
    \label{fig:tts-vs-dirichlet-turn-ratio-perturb}
\end{figure}

\paragraph*{Results} In this part, we show the result of the robustness of the heterogeneous controller against turning ratio perturbations. Figure \ref{fig:tts-vs-dirichlet-turn-ratio-perturb} depicts performance degradation as perturbations intensify. Despite this, even at the highest perturbation level, where the mean of the Dirichlet distribution is equal turning ratios (e.g., 1/3 for a link with 3 turns), the heterogeneous controller still outperforms the homogeneous control. It suggests that while the heterogeneous controller is affected by turning ratio perturbations, its adaptive mechanism still provides a superior performance in managing traffic flow under uncertainty. Therefore, the sensitivity analysis demonstrates a strong robustness of our proposed approach. Note that there is no variability for homogeneous control  in Figure \ref{fig:tts-vs-dirichlet-turn-ratio-perturb} because we fixed the tested scenario to ensure the only source of uncertainty is turning ratio perturbation.

\section{Conclusion and Discussion}
This paper investigates the generalization of multi-hop traffic pressure for heterogeneous perimeter control. The multi-hop traffic pressure provides insight not only into immediate downstream link conditions but also into the more distant traffic conditions. Our customizable long-distance traffic metric, grounded in Markov chain theory, bridges the gap between MFDs and 1-hop pressure where the spatial granularity is either too coarse or too fine. To account for the greater influence of closer traffic links, the multi-hop pressure naturally discounts distant downstream links by the product of turning ratios, interpreted as the multi-step transition probability in the Markov chain. This novel metric, supported by clean mathematical formulation, effectively guides heterogeneous perimeter control. A sensitivity analysis of the heterogeneous perimeter controller based on multi-hop pressure against turning ratio perturbation is conducted.

Analysis of results provides insights into the multi-hop pressure based on the heterogeneous controller. Within the optimal range of pressure sensitivity, the controller performs better than when the pressure sensitivity is beyond the optimal range, which emphasizes the calibration of control parameters. Besides, the performance convergence over hops indicates that knowing extremely distant traffic conditions does not contribute to performance improvement. Such convergence can be explained by the decay of downstream link importance over large hops. Furthermore, our result shows that performance improvement is greater when the demand heterogeneity is larger, whereas under the most homogeneous demand, the improvement is minimal. In addition, to evaluate the robustness of the heterogeneous controller, the sensitivity analysis involves experiments with different levels of perturbation toward turning ratios, and the result demonstrates strong robustness for the heterogeneous controller.

Future works could further 1) generalize the pressure definition by extending the upstream potential to multiple hops, 2) incorporate dynamic turning ratio estimation in multi-pressure calculation, 3) formulate other heterogeneous perimeter controllers, 4) test on a more complex city network, and 5) extend the application of multi-hop pressure to other traffic-related problems such as perimeter identification, congestion pricing, and adaptive traffic signal control.

{\appendix[Scalar Version: Multi-hop Downstream Pressure for a Single Link]\label{app} To gently guide the reader step-by-step, we first review the standard 1-hop traffic pressure, then demonstrate how to extend the 1-hop pressure to 2-hop and higher-hop versions.

\paragraph*{1-hop Pressure} Adopted from physics, the existing standard 1-hop traffic pressure \cite{varaiya2013maxpressure} is defined as the difference between upstream queue density and the summation of 1-hop downstream queue densities weighted by turning ratios. Mathematically, for a  link $l$, its 1-hop pressure is:
\begin{align}
p(l, 1) = Q(l) - \sum_{j \in \mathcal{N}(l,1)} T_{lj} Q(j)
\end{align}
Apparently, 1-hop pressure is myopic in terms of knowing the traffic conditions further down the link of interest. When considering further neighborhoods, the concept of 1-hop pressure can be extended to multi-hop pressure to account for a wider range of traffic networks, capturing the cumulative effect of traffic congestion in neighboring areas.

\paragraph*{2-hop Pressure} As per the requirement for multi-hop pressure, the congestion at 2-hop downstream links has less influence, and is naturally discounted by the turning ratio from link $l$ to 1-hop downstream links:
\begin{align}
    p(l,2) &= Q(l) - \sum_{i_1 \in \mathcal{N}(l,1)} T_{l i_1} \Big( Q(i_1) + \sum_{i_2\in \mathcal{N}(i_1,1)}T_{i_1 i_2}Q(i_2) \Big) \\
    &= \underbrace{Q(l) - \sum_{i_1 \in \mathcal{N}(l,1)} T_{l i_1}Q(i_1)}_{p(l,1)} \nonumber \\ & \phantom{=p(l,1)} - \sum_{i_1\in \mathcal{N}(l,1)}  \sum_{i_2 \in \mathcal{N}(i_1,1)} T_{l i_1} T_{i_1 i_2}Q(i_2) \\
    &= p(l,1) - \sum_{i_1\in \mathcal{N}(l,1)}  \sum_{i_2 \in \mathcal{N}(i_1,1)} T_{l i_1} T_{i_1 i_2}Q(i_2)
\end{align}


\paragraph*{Multi-hop Pressure} 
As you may notice, there exists a recursive relationship between two consecutive hops of pressure. The second term represents the additional pressure exerted on link $l$ due to congestion $h$ hops away that was not accounted for in $(h-1)$-hop pressure. By subtracting the second term, we isolate the pressure that is specifically due to the 
$h$-hop congestion, distinguishing it from the cumulative pressure calculated up to $h-1$ hops:
\begin{multline}
    p(l, h) = p(l, h-1) \\
    - \sum_{i_1 \in \mathcal{N}(l, 1)} \sum_{i_2 \in \mathcal{N}(i_1, 1)} \cdots 
    \sum_{i_h \in \mathcal{N}(i_{h-1}, 1)} \\
    T_{l i_1} T_{i_1 i_2} \cdots T_{i_{h-1} i_h} Q(i_h)
\end{multline}

Explained in graph walk and Markov chain, the product of $h$ turning ratios in the second term is the probability of graph walk with exactly $h$ edges that lead from vertex $l$ to vertex $i_h$, and this probability measures the $h$-step influence of $h$-hop downstream link $i_h$ towards link $l$. The summation in the second term accumulates all the influence of $h$-hop downstream links upon traffic link $l$.}

\appendix[Properties of Multi-hop Downstream Pressure]\label{app:property-pressure} The multi-hop downstream pressure has two properties:
\begin{enumerate}
    \item \textbf{Monotonously decreasing over the hop}:
\begin{align}
    p(l, h) \leq p(l, h-1)  \quad \forall l
\end{align}
    \item \textbf{Bounded range} (Assuming queue density is normalized to $[0, 1]$ by dividing $Q_{\text{max}}$): 
    \begin{align}
    p(l, h) \in [-h, 1] \quad \forall l
    \end{align}
\end{enumerate}

 The upper bound is reached when the link itself is full and all its downstream links do not have queued vehicles. In contrast, the lower bound is reached when the link itself does not have any queues while all downstream links are full. 
 

\textbf{Proposition}: Lower bound of $h$-hop pressure is $-h$:
\begin{proof}

\textbf{Base case} ($h=1$): Using the property that the normalized queue density satisfies $0 \leq Q(l) \leq 1$, we have
\begin{align}
    p(l,1) &= Q(l) - \sum_{j \in \mathcal{N}(l,1)} T_{ij} Q(j) \\ &\geq 0 - \sum_{j \in \mathcal{N}(l,1)} T_{ij}\times 1 \\
    &= -1
\end{align}
    
Hence, it establishes the base case.

\textbf{Inductive step}: Assume the proposition holds for $h-1$, i.e., $p(l,h-1) \geq -(h-1)$. For $h$, the following holds:
\begin{align}
    p(l,h) &= p(l,h-1) - \sum_{i_1 \in \mathcal{N}(l, 1)} \sum_{i_2 \in \mathcal{N}(i_1, 1)} ... \sum_{i_h \in \mathcal{N}(i_{h-1}, 1)} \nonumber \\
    &\phantom{\sum_{i_1 \in \mathcal{N}(l, 1)} \sum_{i_2 \in \mathcal{N}(i_1, 1)}} T_{l i_1} T_{i_1 i_2} ... T_{i_{h-1} i_h} Q(i_h)  \\    
    &\geq -(h-1) - \sum_{i_1 \in \mathcal{N}(l, 1)} \sum_{i_2 \in \mathcal{N}(i_1, 1)} ... \sum_{i_h \in \mathcal{N}(i_{h-1}, 1)} \nonumber \\
    &\phantom{\sum_{i_1 \in \mathcal{N}(l, 1)} \sum_{i_2 \in \mathcal{N}(i_1, 1)}} T_{l i_1} T_{i_1 i_2} ... T_{i_{h-1} i_h} \times 1 \\
    &= -(h-1) - \sum_{i_1 \in \mathcal{N}(l, 1)} T_{l i_1} \sum_{i_2 \in \mathcal{N}(i_1, 1)} T_{i_1 i_2} ... \nonumber \\
    &\phantom{=-(h-1)- \sum_{i_1 \in \mathcal{N}(l, 1)} T_{l i_1}} \sum_{i_h \in \mathcal{N}(i_{h-1}, 1)}  T_{i_{h-1} i_h} \\
    &= -(h-1) - 1 \times 1 ... \times 1 \\
    &= -(h-1) - 1 \\
    &= -h
\end{align}
Thus, by the principle of induction, the lower bound of $h$-hop pressure is $-h$ for all $h \geq 1$.
\end{proof}


\bibliographystyle{IEEEtran}
\bibliography{refs}



\begin{IEEEbiography}[{\includegraphics[width=1in,height=1.25in,clip,keepaspectratio]{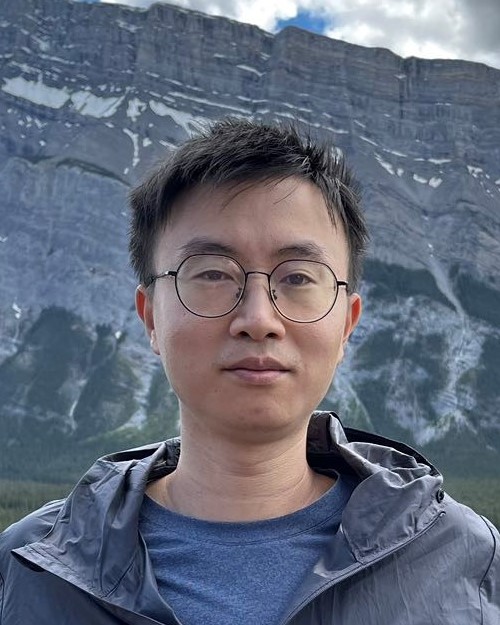}}]{Xiaocan Li}
received the B.Eng. degree from Beihang University, Beijing, China, in 2017, and the M.Sc. degree in Control Theory and Engineering from the Institute of Automation, Chinese Academy of Sciences, Beijing, China, in 2020. He is currently a Ph.D. candidate with the Department of Mechanical \& Industrial Engineering at the University of Toronto, ON, Canada. His research interests include deep reinforcement learning, spatiotemporal prediction, and their application to intelligent transportation systems.
\end{IEEEbiography}

\begin{IEEEbiography}[{\includegraphics[width=1in,height=1.25in,clip,keepaspectratio]{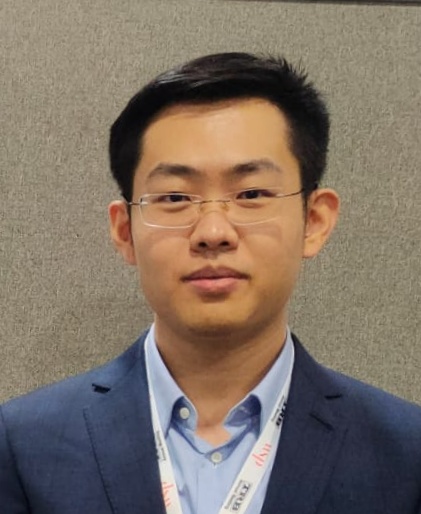}}]{Xiaoyu Wang}
received the B.Eng. degree in automation from Tianjin University, Tianjin, China, in 2016, and the M.Sc. degree in control science and technology from Shanghai Jiao Tong University, Shanghai, China, in 2019. He is currently a Ph.D. candidate with the Department of Civil and Mineral Engineering at the University of Toronto, ON, Canada. His research interests include control, reinforcement learning, and their application to intelligent transportation and multi-agent systems.
\end{IEEEbiography}

\begin{IEEEbiography}[{\includegraphics[width=1in,height=1.25in,clip,keepaspectratio]{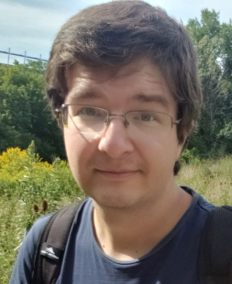}}]{Ilia Smirnov}
is a Research Associate at the Department of Civil and Mineral Engineering at the University of Toronto. He completed a Ph.D. in Pure Mathematics (Algebraic Geometry) at Queen's University, ON, Canada in 2020. His research interests include Planning, Reinforcement Learning, and Optimization in Intelligent Transportation Systems, as well as Enumerative Algebraic Geometry / Intersection Theory.
\end{IEEEbiography}

\begin{IEEEbiography}[{\includegraphics[width=1in,height=1.25in,clip,keepaspectratio]{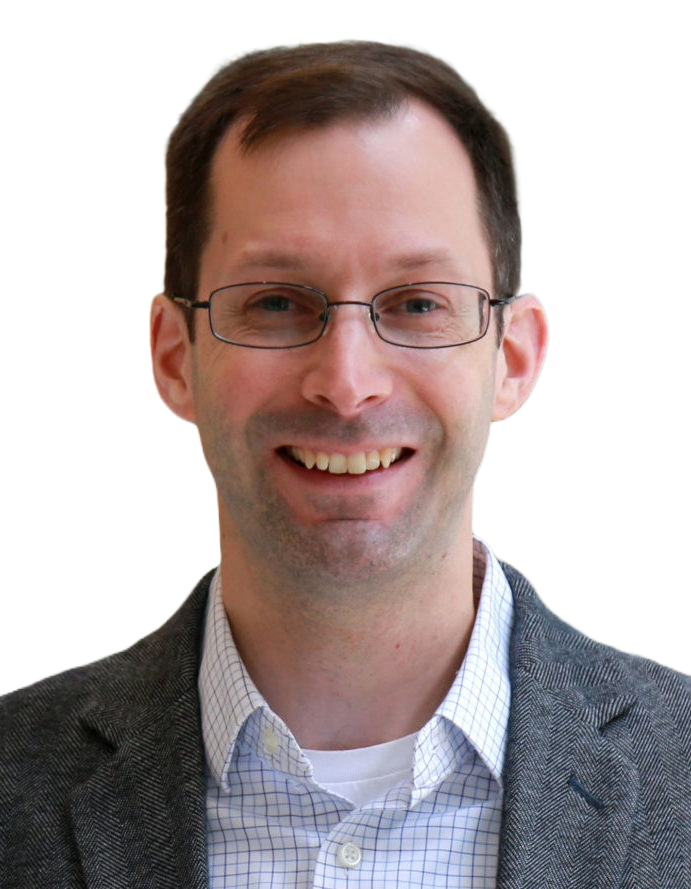}}]{Scott Sanner}
received the B.Sc. degree in computer science from the Carnegie Mellon University, Pittsburgh, PA, USA, in 1999, the M.Sc. degree in computer science from Stanford University, Stanford, CA, USA, in 2002, and the Ph.D. degree in computer science from the University of Toronto, ON, Canada, in 2008. He is an Associate Professor in Industrial Engineering and Cross-appointed in Computer Science at the University of Toronto. Scott’s research focuses on a broad range of AI topics spanning sequential decision-making and applications of machine/deep learning to Smart Cities. Scott is currently an Associate Editor for the Machine Learning Journal (MLJ) and the Journal of Artificial Intelligence Research (JAIR). Scott was a co-recipient of paper awards from the AI Journal (2014), Transport Research Board (2016), CPAIOR (2018) and a recipient of a Google Faculty Research Award in 2020.
\end{IEEEbiography}

\begin{IEEEbiography}[{\includegraphics[width=1in,height=1.25in,clip,keepaspectratio]{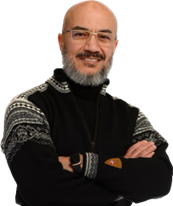}}]{Baher Abdulhai}
received the Ph.D. degree in engineering from the University of California, Irvine, CA, USA, in 1996. He is a Professor in Civi Engineering at the University of Toronto, ON, Canada. He has 35 years of experience in transportation systems engineering and Intelligent Transportation Systems (ITS). He is the founder and Director of the Toronto Intelligent Transportation System Center, and the founder and co-Director of the i-City Center for Automated and Transformative Transportation Systems (iCity-CATTS).
He received several awards including IEEE Outstanding Service Award, Teaching Excellence award, and research awards from Canada Foundation for Innovation, Ontario Research Fund, and Ontario Innovation Trust.  He served on the Board of Directors of the Government of Ontario (GO) Transit Authority from 2004 to 2006.  He served as a Canada Research Chair (CRC) in ITS from 2005 to 2010. His research team won international awards including the International Transportation Forum innovation award in 2010 (Hossam Abdelgawad), IEEE ITS 2013 (Samah El-Tantawy) and INFORMS 2013 (Samah El-Tantawy). In 2015 he has been inducted as a Fellow of the Engineering Institute of Canada (EIC). In 2018, he won the prestigious CSCE Sandford Fleming (Career Achievement) Award for his contribution to transportation in Canada. He has been elected Fellow of the Canadian Academy of Engineering in 2020. In 2021, he won the Ontario Professional Engineers Awards (OPEA) Engineering Medal for career Engineering Excellence.
\end{IEEEbiography}

\end{document}